\documentclass{article}
\usepackage{graphicx} 
\usepackage[skip=0.333\baselineskip]{caption}
\usepackage{subcaption}
\usepackage[utf8]{inputenc}
\usepackage{url}            
\usepackage{booktabs}       
\usepackage{enumitem}
\usepackage{natbib}
\usepackage{amsfonts}       
\usepackage{nicefrac}       
\usepackage{microtype}      
\usepackage{xcolor}         
\usepackage{enumitem}
\usepackage{amsthm}
\usepackage{amsmath,amsfonts,bm}
\makeatletter
\newtheorem*{rep@theorem}{\rep@title}
\newcommand{\newreptheorem}[2]{%
\newenvironment{rep#1}[1]{%
 \def\rep@title{#2 \ref{##1}}%
 \begin{rep@theorem}}%
 {\end{rep@theorem}}}
\makeatother

\newreptheorem{theorem}{Theorem}

\usepackage[framemethod=TikZ]{mdframed}
\mdfdefinestyle{MyFrame}{%
    linecolor=black,
    outerlinewidth=.3pt,
    roundcorner=5pt,
    innertopmargin=1pt, 
    innerbottommargin=1pt, 
    innerrightmargin=1pt,
    innerleftmargin=1pt,
    backgroundcolor=black!0!white}

\mdfdefinestyle{MyFrame2}{%
    linecolor=white,
    outerlinewidth=1pt,
    roundcorner=2pt,
    innertopmargin=\baselineskip,
    innerbottommargin=\baselineskip,
    innerrightmargin=10pt,
    innerleftmargin=10pt,
    backgroundcolor=black!3!white}

\newcommand{\RNum}[1]{\uppercase\expandafter{\romannumeral #1\relax}}

\newcommand{\vertiii}[1]{{\left\vert\kern-0.25ex\left\vert\kern-0.25ex\left\vert #1 
    \right\vert\kern-0.25ex\right\vert\kern-0.25ex\right\vert}}
\newcommand{\vertiiii}[1]{{\vert\kern-0.25ex\vert\kern-0.25ex\vert #1 
    \vert\kern-0.25ex\vert\kern-0.25ex\vert}}

\usepackage{mathtools}

\newcommand{\cut}[1]{}

\newcommand{\removelatexerror}{\let\@latex@error\@gobble}

\def\eqref#1{Eq.~\ref{#1}}

\def\1{\bm{1}}

\DeclareMathAlphabet{\mathsfit}{\encodingdefault}{\sfdefault}{m}{sl}
\SetMathAlphabet{\mathsfit}{bold}{\encodingdefault}{\sfdefault}{bx}{n}

\usepackage{todonotes}
\usepackage{amssymb,fge}
\usepackage{thm-restate}
\usepackage{wrapfig}
\usepackage{bbm}
\usepackage{mathrsfs}
\usepackage{soul}
\usepackage{array}
\usepackage{multirow}
\usepackage{afterpage}
\usepackage{changepage} 
\usepackage{floatpag}
\usepackage{amsmath}
\usepackage{algorithm}
\usepackage{algorithmic}
\usepackage{booktabs}
\usepackage{microtype}
\usepackage{float}
\usepackage{hyperref}
\usepackage{caption}

\usepackage[accepted]{icml2024}

\newcommand{\fakesubsection}[1]{%
  \par\refstepcounter{subsection}
  \subsectionmark{#1}
  \addcontentsline{toc}{subsection}{\protect\numberline{\thesubsection}#1}
}

\usepackage{amsmath}
\usepackage{amssymb}
\usepackage{mathtools}
\usepackage{amsthm}

\date{}
\begin{document}
\twocolumn[
\icmltitle{The Uncanny Valley: A Comprehensive Analysis of Diffusion Models}



\icmlsetsymbol{equal}{*}

\begin{icmlauthorlist}
\icmlauthor{Karam Ghanem}{sch}
\icmlauthor{Danilo Bzdok}{sch}
\end{icmlauthorlist}

\icmlaffiliation{sch}{Mila - Quebec Artificial Intelligence Institute Montreal, Quebec, Canada}

\icmlcorrespondingauthor{Karam Ghanem}{ghanemkaram12@gmail.com}
\icmlcorrespondingauthor{Danilo Bzdok}{danilobzdok@gmail.com}

\icmlkeywords{Machine Learning, ICML}

\vskip 0.3in
]

\printAffiliationsAndNotice{} 

\begin{abstract}

Through Diffusion Models (DMs), we have made significant advances in generating high-quality images. Our exploration of these models delves deeply into their core operational principles by systematically investigating key aspects across various DM architectures: i) noise schedules, ii) samplers, and iii) guidance. Our comprehensive examination of these models
sheds light on their hidden fundamental mechanisms, revealing the concealed foundational elements that are essential for their effectiveness. Our analyses emphasize the hidden key factors that determine model performance, offering insights that contribute to the advancement of DMs. Past findings show that the configuration of noise schedules, samplers, and guidance is vital to the quality of generated images; however, models reach a stable level of quality across different configurations at a remarkably similar point, revealing that the decisive factors for optimal performance predominantly reside in the diffusion process dynamics and the structural design of the model's network, rather than the specifics of configuration details. Our comparative analysis reveals that Denoising Diffusion Probabilistic Model (DDPM)-based diffusion dynamics consistently outperform the Noise Conditioned Score Network (NCSN)-based ones, not only when evaluated in their original forms but also when continuous through Stochastic Differential Equation (SDE)-based implementations.

\end{abstract}

\section{Introduction}

The landscape of machine learning is perpetually evolving, with Diffusion Models (DMs) emerging as a leader in image generation \cite{kazerouni2023diffusion}. As DMs become increasingly complex, recent studies have attempted to summarize the mechanisms behind their function, which is crucial for theoretical comprehension and practical application \cite{chang2023design, yang2023diffusion, Croitoru_2023}. While past studies have provided important insights, they lacked in effectively comparing DMs for proportional scaling, a key aspect for further progress in the field. The metrics presently employed to assess generative models raise concerns regarding their comparability, due to inconsistent application, and the substantive significance of the scores they produce \cite{betzalel2022study}. Identifying the key aspects of DMs that warrant further research necessitates more than a summary of current functionality but a comprehensive examination of every facet of these models.

Our study has delved deeper, conducting rigorous empirical analyses to identify and analyze the primary driving forces behind performance in the DM landscape. We have built on the available quality metrics to create experimental scenarios that are optimized towards direct comparability across DM variants, which sheds light on the fundamental DM mechanics. Our current research has examined the progression of Denoising Diffusion Probabilistic Models (DDPMs) as they integrate with Noise Conditioned Score Networks (NCSNs), culminating into the advanced development of Stochastic Differential Equation (SDE)-based models, thus providing an understanding of their intricate dynamics and interplay \cite{song2021scorebased} throughout the DM evolution process. By focusing on the fundamentals that underpin state of the art (SOTA) DMs, we have charted a path for future exploration, aiming to optimize the comparability between, and output quality of these models.

The fundamentals we explored have proven to be building blocks in the field, serving as the basis for a multitude of recently proposed variants and adaptations within the domain. For example, we have carefully benchmarked the Denoising Diffusion Implicit Model (DDIM)
sampler, whose structure has been used to create a number of ODE solvers for sampling from SDE-based DMs \cite{liu2022pseudo,karras2022elucidating,lu2023dpmsolver,zhao2023unipc}. Meanwhile, traditional Markovian and Langevin samplers from DDPMs and NCSNs respectively, can be seen as components of the broader predictor-corrector sampling framework \cite{song2021scorebased,allgower2012numerical, lezama2023discrete} and hence we analyzed them separately. We have also analyzed beyond the fundamentals, by examining a number of samplers for additional comparability, exploring the realm of noise schedules thoroughly, and examining Classifier Guidance through an ablation analysis. Our study reveals that the effectiveness of DMs is significantly influenced by the diffusion dynamics and network design, with DDPM-based diffusion dynamics demonstrating the best performance.

\section{Background}
\subsection{NCSNs}

NCSNs approximate the score of the data distribution, $\nabla_x \log p_{\text{data}}(x)$, using a neural network $s_{\theta}$. The network is trained with denoising score matching to estimate scores across multiple noise levels without requiring computation of higher-order gradients \cite{JMLR:v6:hyvarinen05a}. The training objective minimizes the expected squared difference between the network's output and the true data score. Samples from the data distribution are generated using annealed Langevin dynamics. This process involves starting with a prior distribution and progressively refining the samples through Langevin steps across decreasing noise scales \cite{song2020generative, song2020improved}.

\subsection{DDPMs}
DDPMs are a class of latent variable models that generate data by reversing a diffusion process. The forward process is a fixed Markov chain that gradually adds Gaussian noise to the data. The noise level is controlled by a variance schedule $\beta$ (More on Noise Schedules in \hyperref[sec:background_appendix]{Appendix}). The reverse process is a learned Markov chain that aims to denoise the data, effectively reconstructing the original data distribution from the noised data \cite{ho2020denoising, sohldickstein2015deep}.  The forward and reverse processes are shown in Figure~\ref{fig:Model_Evolution_Chart} where $x_0$ is the original data, $T$ is the total number of diffusion steps. Training involves optimizing the variational lower bound on the negative log likelihood $\mathbb{E}_q \left[ -\log \frac{p_{\theta}(x_{0:T})}{q(x_{1:T}|x_0)} \right]$. The reverse process of DDPMs is parameterized to estimate the mean and variance of the Gaussian transitions in the reverse Markov chain. The loss function used for training is a simplified variational bound that emphasizes the reconstruction of data from noisier states with larger timesteps in the diffusion process. The sampling process reverses the diffusion process, starting from noise and progressively denoising it to generate a sample \cite{ho2020denoising}.

\subsection{DDIM}

DDIMs offer an advanced approach in DM sampling, leveraging non-Markovian forward processes for more efficient generative modeling. This is achieved by defining a family of inference distributions parameterized by $\sigma$, which determines the stochasticity of the forward process. The generative model, trained via variational inference, capitalizes on the predicted clean data to sample the subsequent noisier data. This model is supposed to be able to operate under fewer iterations than Markovian sampling by employing a fixed surrogate objective function \cite{song2022denoising}.

\subsection{SDE-based Diffusion}

\begin{figure*}[htbp!]

    \begin{subfigure}{0.98\textwidth}
        \centering
        \includegraphics[width=\textwidth]{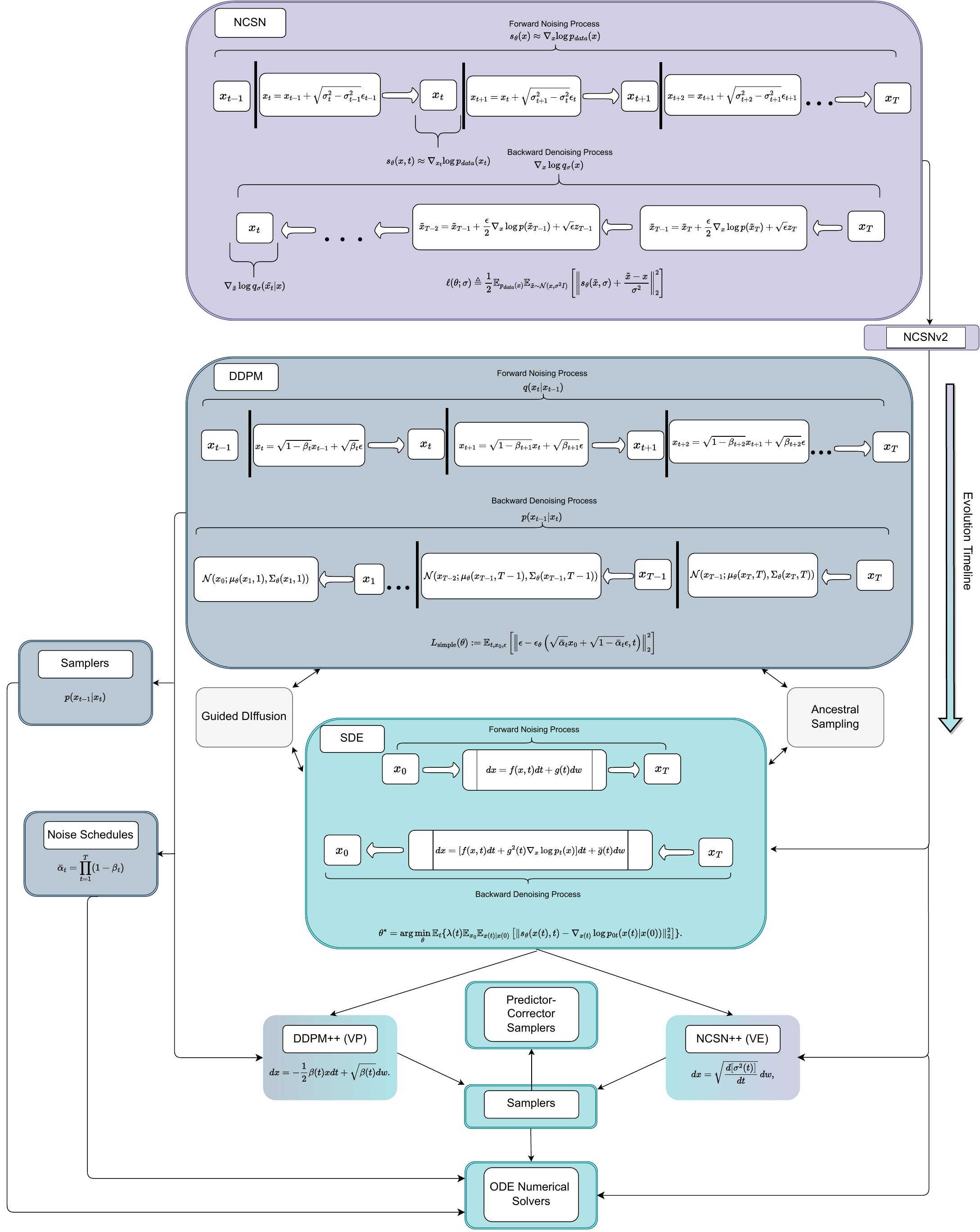}
        \clearpage
    \end{subfigure}
    \caption{Evolutionary trajectory of DMs, from the progression of NCSNs (purple) through DDPMs (gray) to the more recent SDE-based models (cyan). The upper segment illustrates the NCSN's forward and backward processes with variable noise levels , while the middle segment details the DDPM's noising and denoising mechanics. The lower segment delineates the SDE approach, showing SDE variants.}
    \label{fig:Model_Evolution_Chart}

\end{figure*}

SDE-based diffusion generalizes the idea of perturbing data with noise, transitioning from a data distribution \( p_0 \) to a tractable prior \( p_T \) via an infinite number of noise scales, characterized by a continuous-time variable \( t \in [0, T] \). The diffusion process is defined by an ODE, as shown in Figure~\ref{fig:Model_Evolution_Chart} where \( f(x, t) \) represents the drift coefficient, \( g(t) \) the diffusion coefficient, and \( w \) the standard Wiener process. The goal is to map data to an unstructured noise distribution (the prior) using the SDE and then reverse this process for sample generation \cite{song2021scorebased}. By reversing the SDE, samples of the prior \( p_T \) are initiated to generate samples of the original data distribution \( p_0 \) \cite{anderson1982reverse}. The reverse-time SDE is shown in Figure~\ref{fig:Model_Evolution_Chart}
where \( \nabla_x \log p_t(x) \) is the score of the marginal distribution at time \( t \).
\( \nabla_x \log p_t(x) \) is estimated by training a time-dependent score-based model \( s_\theta(x, t) \) using a continuous generalization of score matching
where \( \lambda(t) \) is a time-dependent positive weighting function \cite{JMLR:v6:hyvarinen05a, song2020generative}. The two primary types of SDEs are Variance Exploding (VE) associated with NCSNs and Variance Preserving (VP) associated with DDPMs. VE SDEs exhibit an increasing variance with time and are defined as $dx = \sqrt{\frac{d}{dt} \sigma^2(t)} dw$, where $\sigma(t)$ is the noise scale function. VP SDEs are formulated as $dx = -\frac{1}{2} \beta(t) x dt + \sqrt{\beta(t)} dw$, with $\beta(t)$ controlling the noise level, maintaining constant variance over time. For sample generation from SDEs numerical ODE solvers, such as the Euler-Maruyama method, help approximate SDE trajectories, enabling the generation of samples by discretizing the reverse-time SDE in a manner consistent with the forward dynamics\cite{kloeden2013numerical, song2021scorebased}.

\subsection{Classifier Guidance}

Classifier Guidance leverages a pre-trained classifier, denoted as \( p_{\phi}(y|x_t, t) \), to direct the generative process towards specific class labels. The classifier is trained on noisy images, where classifier's gradients are utilized, \( \nabla_{x_t} \log p_{\phi}(y|x_t, t) \), to steer the diffusion sampling towards a desired label \( y \) \cite{dhariwal2021diffusion}. This is executed by adjusting the reverse noising process, where each diffusion step is sampled from an approximation of the conditional distribution \( p_{\theta,\phi}(x_t|x_{t+1}, y) \), effectively shifting the Gaussian transition's mean by the scaled gradient, thus facilitating the generation of class-specific samples.

\fakesubsection{}

\clearpage 
\clearpage

\section{Methods}\label{sec:methods}

    \subsection{Quality Metrics: Optimizing for Comparability}

 In the realm of generative models in general and DMs in particular, the assessment of sample quality plays a pivotal role. Our study embarks on an in-depth performance analysis, chiefly concerned with the reliability and comparability of sample quality metrics across DMs and how to work around the weaknesses of these metrics. A significant challenge encountered in our comparative analyses is the unreliability of Frechet Inception Distance (FID). The dependence of FID on the size of the generated dataset has often led to inconsistent practices, where different studies have resorted to varying generated sample set sizes for FID calculations \cite{song2020generative, song2020improved, borji2021pros}. This variability can dramatically influence FID, affecting the ability to compare results consistently. Furthermore, FID is heavily contingent upon the reference dataset. Variations in the choice or subsets of the reference dataset used across studies compound the difficulty in establishing a common ground for comparison \cite{dhariwal2021diffusion}.

On the other hand, Inception Score (IS) offers a more stable metric for comparison, thanks to the consistency provided by the pre-trained Inception v3 model used in its computation \cite{barratt2018note}. However, just like FID, the applicability of the IS for comparative purposes is constrained to models trained on the same dataset. This restriction arises from the inherent bias of the IS towards ImageNet, upon which Inception v3 was trained \cite{borji2021pros}. As a result, each dataset has a unique IS range when used to train DMs. Despite this, the biases introduced by the IS are uniform across the research community, given that Inception v3 serves as an established standard for score calculation. While IS does offer consistency, its reliability as a metric is limited to models trained on datasets that have the similar classes to ImageNet \cite{borji2021pros,barratt2018note,kynkäänniemi2019improved}. Hence, its reliability is still limited to the nature of the dataset classes. Nonetheless training the models on datasets that have similar classes to ImageNet such as CIFAR10 under the same training conditions and using IS for evaluation under different stages of training, could possibly provide the most reliant approach to exhaustively viewing generative models with comparability. 

The robustness of the IS is further evidenced by its relative insensitivity to the number of generated samples under evaluation. Our empirical investigations reveal that the number of generated samples is overall less impactful to IS compared to FID and especially after going above 10000 generated samples, as shown by Figure~\ref{fig:IS_mean_std}. This finding stands in stark contrast to the conventional wisdom that advocates for the generation of 50000 images to ensure a reliable FID value \cite{borji2021pros,barratt2018note,kynkäänniemi2019improved}. Even with such an established convention, past studies have varied widely in their choice of generated sample sizes for FID calculations, with some using as few as a thousand samples \cite{song2020generative, song2020improved}, and have also varied with the resulting FID score ranges \cite{ho2020denoising, jabri2023scalable, song2022denoising, dhariwal2021diffusion, bao2022analyticdpm, karras2022elucidating, song2021scorebased} which diminishes comparability between studies. Although a past study conveys that FID should achieve stability at 50000 generated images \cite{kynkäänniemi2019improved}, we find that the FID score adopts incomparable ranges of values at different generated sample set sizes with no indicator that the score range stabilizes at 50000 generated images, indicating too high of a correlation to the generated sample set size. Our evidence receives further support by the ranges of FID scores reported by various studies that use SOTA models, which shows the lack of consistency \cite{zhao2023unipc, ho2020denoising, jabri2023scalable, song2022denoising, dhariwal2021diffusion, bao2022analyticdpm, karras2022elucidating, song2020generative, song2020improved, borji2021pros, chen2023importance}. Another core issue is that a low FID score does not necessarily reflect the true generative capability of the model given that aiming for the lowest FID allows the model to be vulnerable to overfitting \cite{jiralerspong2023feature}. In contrast to FID, we confirm that IS yields minimal variance between generated sample sizes after a certain threshold, as shown by Figure~\ref{fig:IS_mean_std}. Unlike our findings with FID, we find that using a generated sample size of 10000 images for IS is sufficient. Furthermore, the IS score range has been more consistent in past studies and hence we find that the IS is more valuable for our purpose of optimizing for comparability of model performance \cite{ho2020denoising, song2020generative, 
 song2020improved, zhao2023unipc, song2021scorebased, chen2023importance}.

Through our meticulous analysis, we aim to understand the effects of the variants of DMs which are deemed to be foundational to DM theory. Our analysis aims not to pursue the best possible metric scores but to facilitate meaningful comparisons between models at different stages of training. This goal informs our choice to report the IS as a function of training epochs for DM variants using the CIFAR10 dataset. We hold training hyperparameters and conditions constant throughout all experiments when possible, and we test the effects of different fundamental DM components on performance by varying the component under question. We also trained the same models on FFHQ and sampled from them to further observe the behavior of these models by visually evaluating the quality of the sampled images. We find that IS does not yield insightful results with FFHQ, given that FFHQ is an image dataset containing high-quality human face images without any ImageNet classes.

 It is not regular practice to evaluate generative models at different intervals of training as we have in our in-depth analysis, given that using large sample sets of generated images for model performance evaluation carries computational and time constraints. The lack of rigorous evaluation within the community has affected the ability to effectively train and apply different forms of DMs. Furthermore, this shortfall in thorough assessment has impeded the community's comprehension of the importance of past research in this domain.  Past studies have also regularly attempted to train models for extensive periods of time to achieve superior FID/IS values. Such practice does not allow for model comparability and is questionable, as it is causing a cascaded issue in understanding the value of the results yielded by studies that are meant to improve the capabilities of generative models. Conversely, various studies have been conducted to improve the established performance metrics which take into account their weaknesses and flaws \cite{borji2021pros,jiralerspong2023feature, kynkäänniemi2019improved}, but the generative model community does not seem to be willing to move away from the traditional metrics due to the very same issue we have tried to work around through our in-depth analysis: lack of comparability and consistency.  

\subsection{Model Convergence from the lens of a Metric}

To enhance the comparability of performance between different DMs, we need to juxtapose the results of the different DMs at their relative points of convergence to evaluate their contributions, hence defining convergence in DMs necessitates a nuanced approach. Traditional models often correlate convergence with a singular loss metric. In contrast, DMs demand a more comprehensive strategy that combines quantitative metrics with qualitative evaluations. The convergence of DMs, in this study, is inferred from the behavior of the IS across epochs. A plateau in IS as a function of epochs indicates that the model has reached a state of equilibrium in its learning process, where additional training does not markedly enhance the model performance. This plateau could indicate either peak learning or a risk of overfitting, potentially reducing image diversity and lowering the IS.

\subsection{Misguided Diffusion}

Understanding the sensitivity of the DM variants to guidance, can help us recognize which of the DM tools under question in our study are prone to guidance and which have the most potential in assisting in learning the training dataset without overfitting. Guided Diffusion in DMs is typically employed to steer the generative process, aiming to create samples that align with specific classes or attributes \cite{dhariwal2021diffusion, ho2022classifierfree}. Guided Diffusion enhances the model's capability to produce high-quality, diverse, and class-specific images, imbuing the generative process with semantic constraints that yield meaningful and task-relevant outputs \cite{dhariwal2021diffusion, ho2022classifierfree}.

\begin{center}
\begin{algorithm} 
\caption{Misguided Diffusion, given a trained DM and an untrained classifier $p_{\phi}(y|x_t)$}
\begin{algorithmic}[1]
\INPUT{class label $y$, sampler type}
\STATE $x_T \gets$ sample from $\mathcal{N}(0, \mathbf{I})$
\FOR{$t = T$ to $1$}
    \IF{sampler type is Markovian}
        \STATE $\mu, \Sigma \gets \mu_{\theta}(x_t), \Sigma_{\theta}(x_t)$
        \STATE $x_{t-1} \gets$ sample from $\mathcal{N}(\mu + \Sigma \nabla_{x_t} \log p_{\phi}(y|x_t),$ $\Sigma)$
    \ELSIF{sampler type is DDIM}
        \STATE $\hat{\epsilon} \gets \epsilon_{\theta}(x_t) - \sqrt{1 - \bar{\alpha}_t} \nabla_{x_t} \log p_{\phi}(y|x_t)$
        \STATE $x_{t-1} \gets \sqrt{\bar{\alpha}_{t-1}} \left( \frac{x_t - \sqrt{1-\bar{\alpha}_t}\hat{\epsilon}}{\sqrt{\bar{\alpha}_t}} \right) + \sqrt{1 - \bar{\alpha}_{t-1}}\hat{\epsilon}$
    \ENDIF
\ENDFOR
\STATE $x_0$
\end{algorithmic}
\end{algorithm}

\end{center}

 We conduct an ablation analysis using "Misguided Diffusion" by utilizing an untrained classifier to influence the sampling process. This reversal of the standard approach is expected to lead to unpredictable and potentially adverse outcomes. An untrained classifier, devoid of the necessary knowledge to provide accurate guidance, is likely to offer poor direction, disrupting the sampling process and introducing instability. The lack of a trained reference frame to distinguish between quality outputs and undesirable ones is hypothesized to result in  nonsensical guidance. This increased stochasticity is anticipated to manifest in the generation of non-specific images. Without the classifier's ability to impart class-specific guidance, the resulting images may not correspond to discernible classes, instead embodying attributes from multiple classes or no class at all. Consequently, the overall performance of the model in generating class-specific imagery is projected to decline, potentially yielding outputs lacking coherence and recognizable features, resembling abstract patterns more than clear objects or classes which tests the robustness of the sampler and the reliability of the trained DM.

Through such an analysis, we aim to understand Guided Diffusion through Classifier Guidance. We do so by comparing the performance of different DM noise schedule-sampler configurations without any guidance, under Misguided Diffusion, and under Classifier Guidance. By implementing Misguided Diffusion, we intend to explore how classifier-guided gradients influence the results of the generative process, hence helping us understand the robustness of the generative process further. The classifier loss function, designed to compute gradients of the output relative to the input image, plays a pivotal role in the denoising step of the DM. If the classifier is untrained, the gradients it provides, based on random weights, would not correlate with any meaningful features of the images which will misguide the sampling process.

\section{Results}

In the pursuit of characterizing the performance scaling of DMs' performance, our exploration has led to several key observations that inform the direction of model development and training. The effectiveness of DMs has been observed to be contingent upon the appropriate orchestration of the noise schedule and the sampler employed. This intricate interplay is crucial as it underscores the significance of the noise schedule-sampler combination in the overall performance of the models. 

\begin{figure}[H]
    \centering
    \begin{subfigure}{0.5\textwidth}
        \centering
        \includegraphics[width=\textwidth]{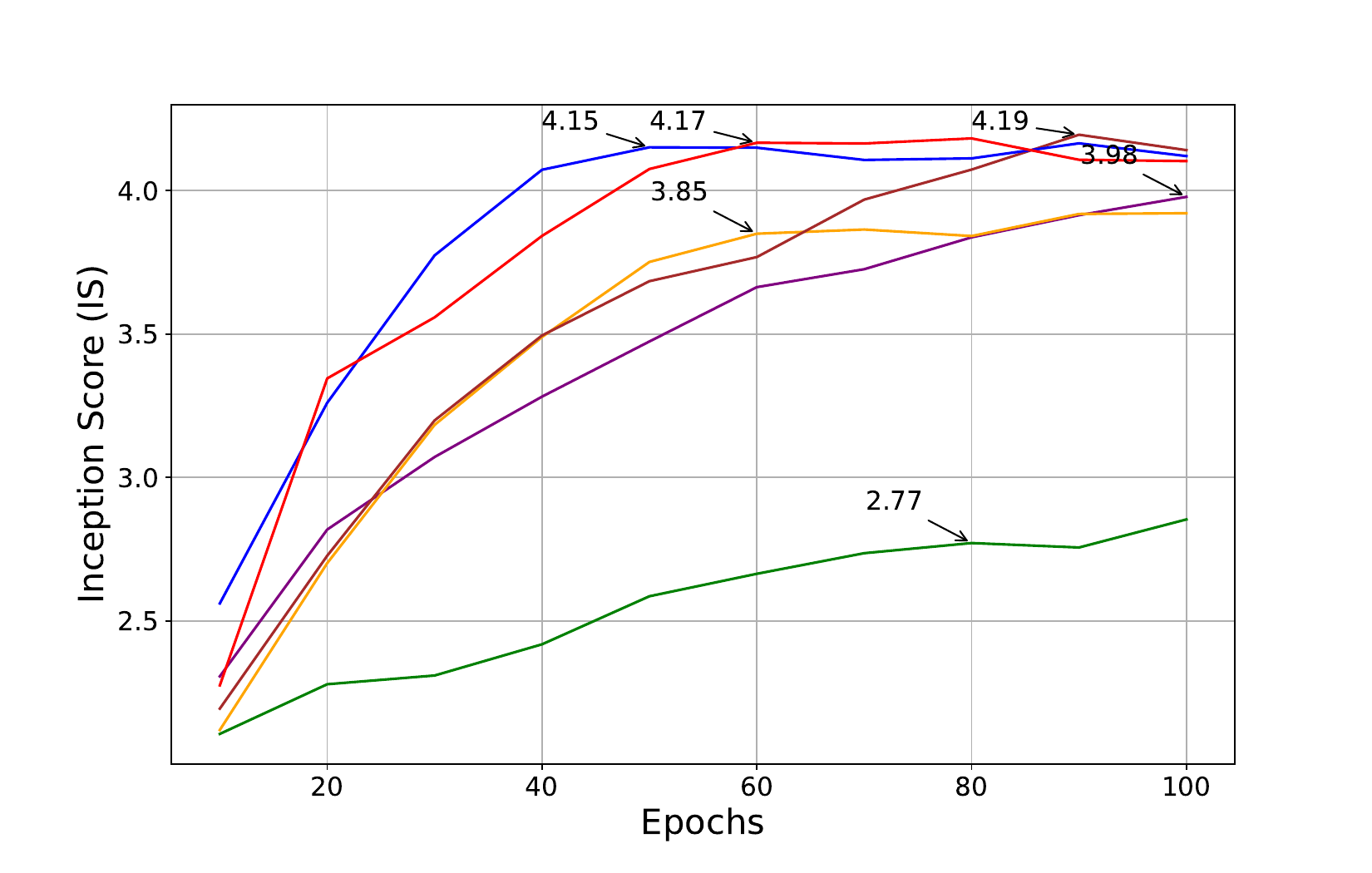}
        \subcaption{}
        \label{subfig:ddim_sampler}
    \end{subfigure}
    \hfill 
    \begin{subfigure}{0.5\textwidth}
        \centering
        \includegraphics[width=\textwidth]{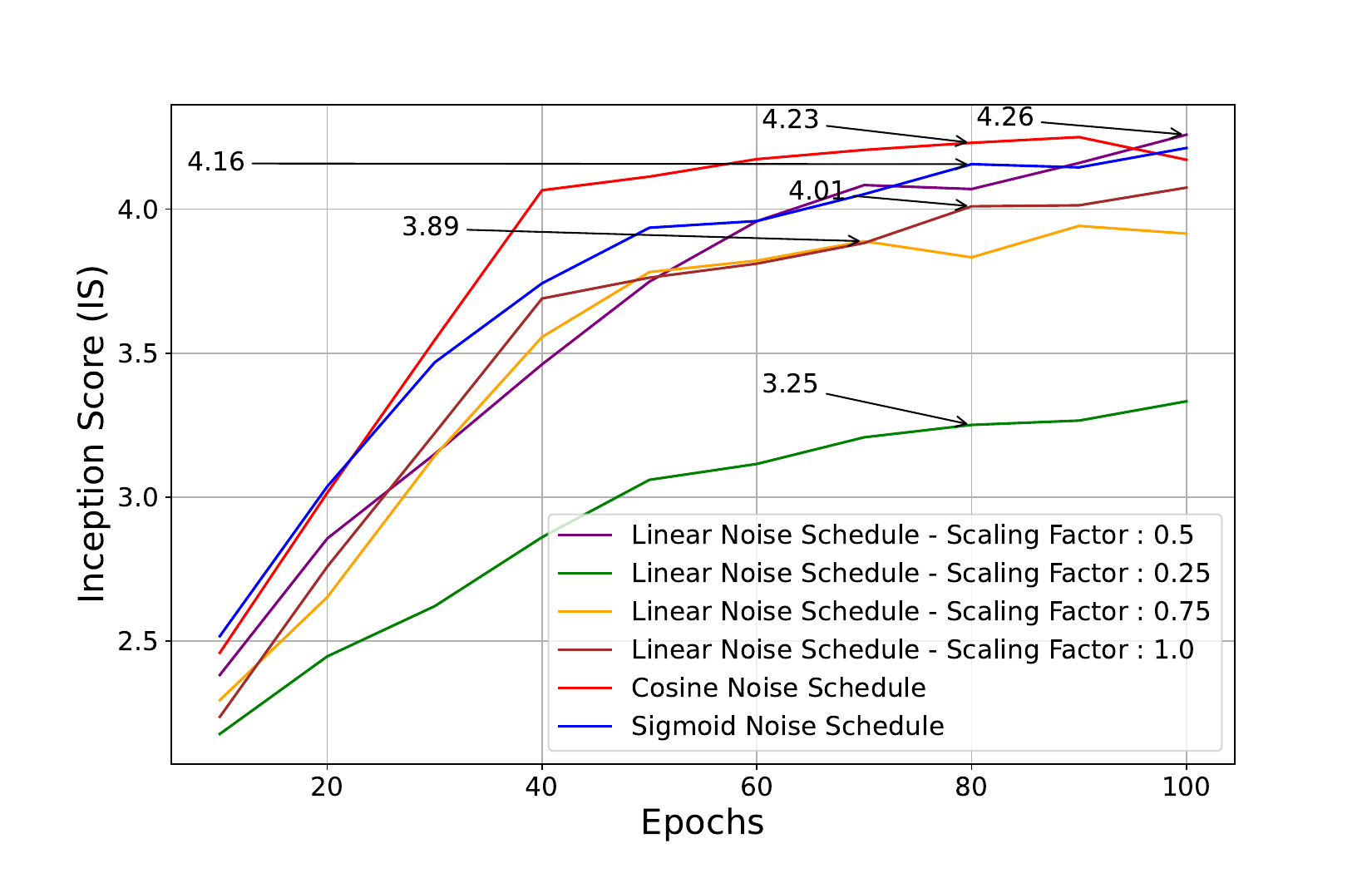}
        \subcaption{}
        \label{subfig:markovian_sampler}
    \end{subfigure}
    \caption{IS as a function of epochs of DDPM variants trained on CIFAR10, using different noise schedules and sampled with a) the DDIM and b) Markovian  samplers.}
    \label{fig:IS_DDPM}
\end{figure}

\paragraph{DDPMs} The comparative performance analysis within the DDPM framework unveiled notable findings when examining the DDIM sampler versus the Markovian sampler. Despite the DDIM sampler's reputation for its strength at generating high-caliber samples expeditiously, our observations suggest that the Markovian sampler stands on equal footing in terms of performance when constrained to a similar number of sampling steps. Nonetheless, the DDIM sampler distinctly exhibits its capabilities by catalyzing model convergence at markedly lower epochs, with discernible enhancements in performance materializing between 50 to 60 epochs for certain noise schedules, as depicted in Figure~\ref{fig:IS_DDPM}. This contrasts with the Markovian sampler's tendency to not reach a similar state of convergence until approximately 80 epochs on average. Our exploration of noise schedules has further elucidated the subtleties in optimizing DDPMs. The cosine and sigmoid noise schedules are exemplary in terms of average performance. The cosine noise schedule, when paired with the Markovian Sampler, presents an expedited convergence within a range of 60-80 epochs, striking an optimal balance between computational resource expenditure and the caliber of the generated output. In a parallel vein, the sigmoid noise schedule, in concert with the DDIM sampler, mirrors this optimal equilibrium, converging within a remarkably brief duration of 40-50 epochs. A key takeaway from our analysis is the discernible underperformance associated with linear noise schedules adopting a scaling factor lower than 0.5. Conversely, linear schedules with scaling factors of exactly 1 and 0.5 are distinguished by their efficacy. A scaling factor of 1 prevails in scenarios devoid of Classifier Guidance with DDIM sampling, while a factor of 0.5 generally emerges as the superior option. Using scaling factors of 1 and 0.5 exhibits a promising potential for enhanced performance with extended training beyond the 100-epoch mark, suggested by the absence of a plateau in the performance curves.

\paragraph{Visual Analysis} When visually inspecting the sampled images from the models trained on CIFAR10, we find that the IS happens to be consistent with the quality of the generated images. Given that FFHQ is not compatible with the IS for reasons mentioned in the \hyperref[sec:methods]{Methods}, we relied on 

visually inspecting the performance of the models. The generated images show that even though the models were able to capture the features of the training set, they were not able to converge even after 200 epochs, as the generated images portray faces of incongruously mixed features that feel unsettling to look at, as shown in Figures \ref{fig:DDPM_FFHQ_80}, \ref{fig:DDPM_FFHQ_150} and \ref{fig:DDPM_FFHQ_200}. On the other hand, while the generated CIFAR10 images are not perfect, as shown in \ref{fig:DDPM_CIFAR10_80}, they feel natural and almost indistinguishable in quality from the CIFAR10 dataset. Such findings imply that while the same network bottleneck resolution was achieved for both datasets, a different network design is necessary to capture the full extent of the complexity of the features of datasets with higher resolutions. This becomes more evident when comparing the images generated by the SDE DMs in Figure~\ref{fig:SDE_NCSN_10_epoch_CIFAR10} to the images generated by DDPMs at convergence in Figure~\ref{fig:DDPM_CIFAR10_80}, as training a non-class conditional SDE VE DM from Figure~\ref{fig:Eulers_method} to 10 epochs, where it is able to achieve the same IS as a converged cosine schedule DDPM model trained to 80 epochs from Figure~\ref{subfig:ddim_sampler}, does not suffice. As shown in  

\begin{figure}[H]
    \centering
    \begin{subfigure}{0.5\textwidth}
        \centering
        \includegraphics[width=\textwidth]{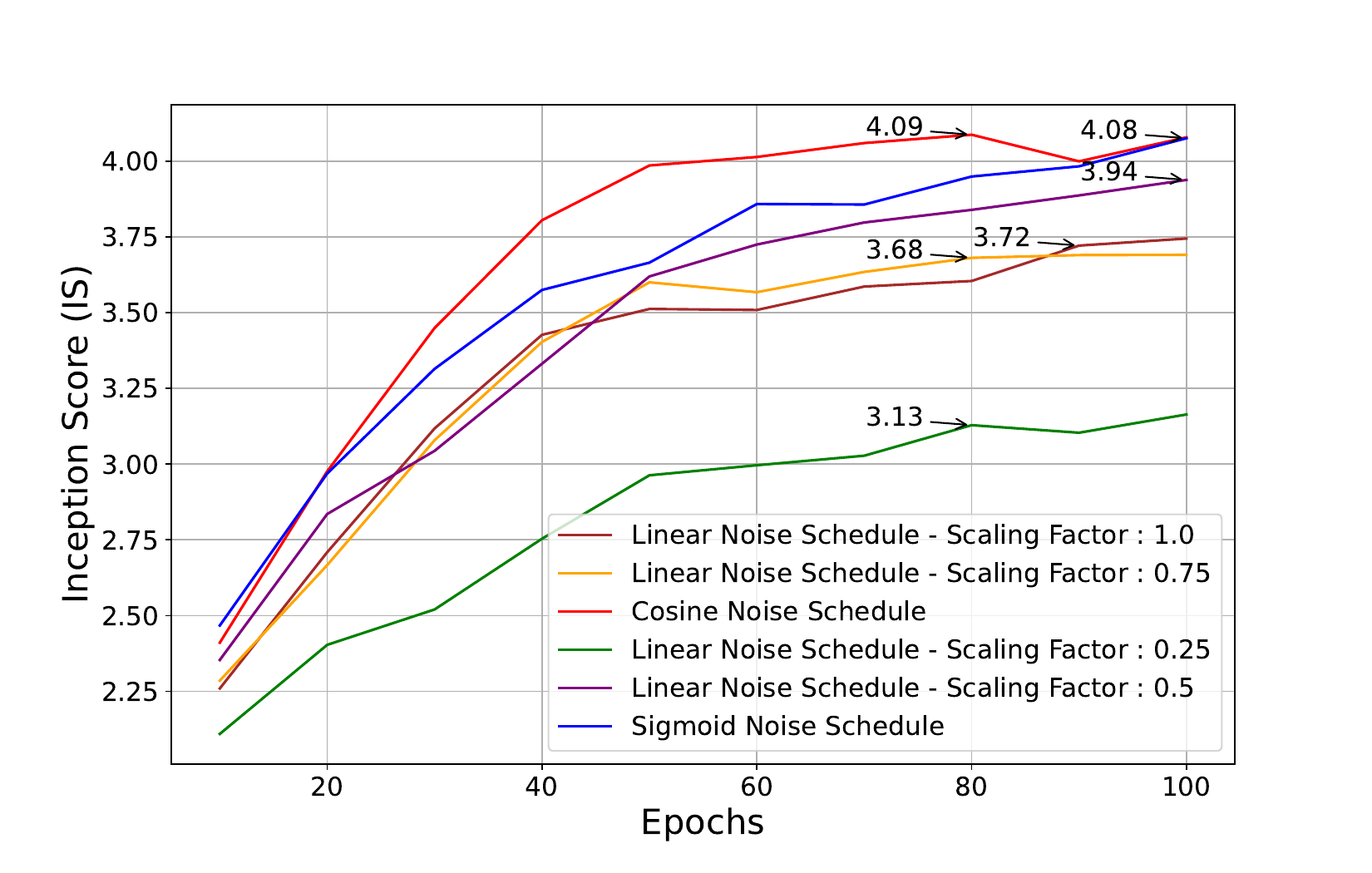}
        \caption{}
        \label{fig:Guidance_1}
    \end{subfigure}
    \hfill 
    \begin{subfigure}{0.5\textwidth}
        \centering
        \includegraphics[width=\textwidth]{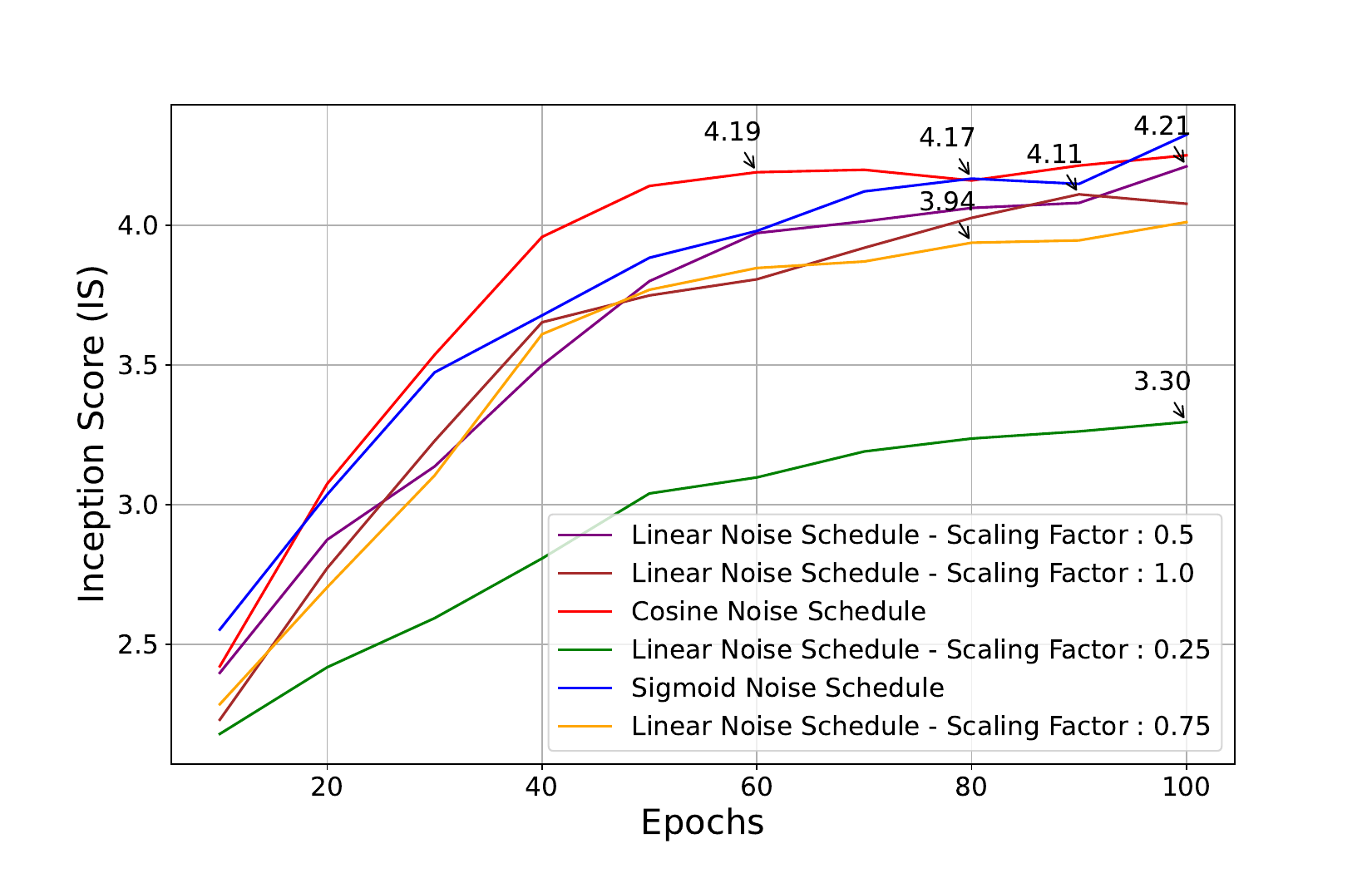}
        \caption{}
        \label{fig:Misguided_1} 
    \end{subfigure}
    \caption{IS as a function of epochs of DDPM variants trained on CIFAR10, using different noise schedules and sampled with the DDIM sampler under a) Classifier Guidance and b) Misguided Diffusion.}
    \label{fig:IS_Guidance}
\end{figure}

 Figure~\ref{fig:SDE_NCSN_100_epoch_CIFAR10}, the SDE model must be pushed closer to its own unique convergence point for the learned features to be generated naturally. This shows that using a measuring metric by itself is not enough to evaluate the state of the model, as while the IS evaluates image quality and image diversity, it fails at understanding when the features of an image are generated so that the image 'feels' coherent. This is further evident when comparing the sampled images from the SDE VP DM at 20 epochs and at 100 epochs in Figures \ref{fig:SDE_DDPM_FFHQ_20} and \ref{fig:SDE_DDPM_FFHQ_100}.

\paragraph{Misguided Diffusion} Our investigations with Misguided Diffusion further reveal that increasing model size is a pivotal factor in elevating performance. The breadth of the model's architecture, endowed with an extensive array of parameters, emerge as the preeminent factors in harnessing the full potential of DMs. This expansion in model size and complexity is paramount in capturing the intricate nuances intrinsic to the the datasets, thus significantly boosting performance. Contrary to conventional expectations, the integration of Classifier Guidance within the model framework did not yield an enhancement in performance. This phenomenon is exemplified in our Misguided Diffusion analysis, which elucidates that irrespective of the efficacy of the training and sampling methodologies, the incorporation of classifier gradients to direct the sampling process neither benefits Classifier Guidance nor impedes the models under Misguided Diffusion, as shown by Figures \ref{fig:IS_Guidance} and \ref{fig:markovian_misguided}. This finding is particularly enlightening as it shows that while ancillary mechanisms like Classifier Guidance may serve niche purposes, such as coaxing the model to generate images from a homogeneous class \cite{dhariwal2021diffusion}, our study shows that they do not inherently enhance the quality of the generated images in general. This distinction emphasizes the role of such tools as situational aides rather than fundamental performance enhancers. Concurrently, the overarching influence of the network design on the training process is evident.  This assertion is corroborated by seminal works that have modified the network design with the nature of the diffusion process \cite{song2021scorebased,ho2020denoising,song2020improved} , where it is inherently shown that considering both the nature of the diffusion process and the network design is crucial as determinants of performance metrics.

\begin{figure}[H]
    \begin{center}
    \begin{subfigure}{0.5\textwidth}
        \centering
        \includegraphics[width=\textwidth]{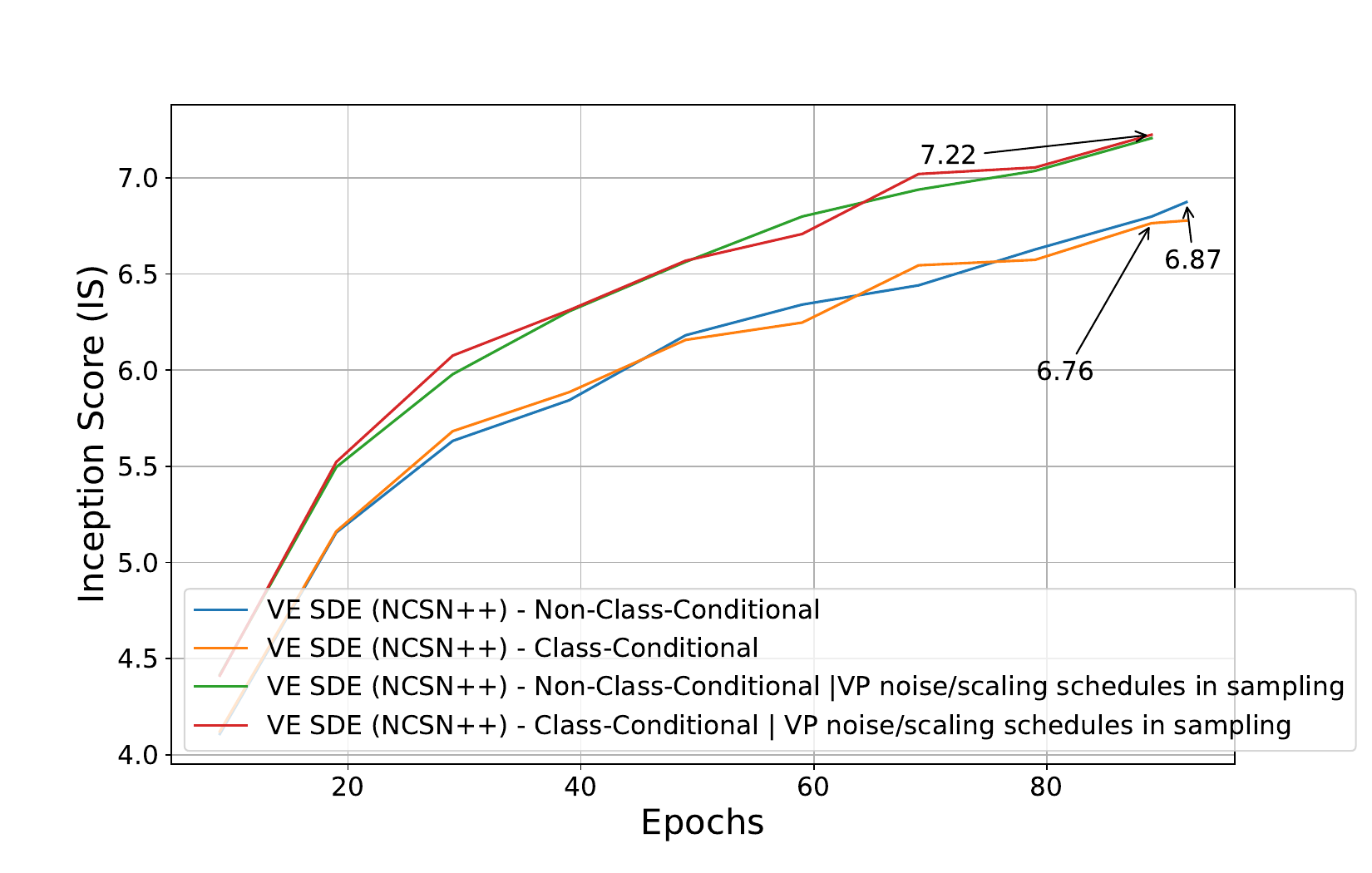}
    \end{subfigure}
    \caption{IS as a function of epochs of of VE SDE DMs trained on CIFAR10 with and without class conditioning, and sampled with the Karras et al. (\citeyear{karras2022elucidating}) reimplemented Euler's ODE solver using both VE noise schedule/scaling schedule parameters and VP noise schedule/scaling parameters.}
    \label{fig:Eulers_VE_VP_parameters}
    \end{center}
\end{figure}

\paragraph{SDE-based DMs} In the pursuit of characterizing the performance scaling of DMs' performance, applying our comprehensive analysis in the realm of SDE DMs has highlighted the following high level points: i) DDPM-based models are frontrunners compared to VE and ii) applying class conditioning has little impact on performance. The first observation is particularly salient when we non-traditionally utilize the VP parameters in Table~\ref{table:Sampling_parameters}, 
 within the Karras et al. re-implemented Euler's method (\citeyear{karras2022elucidating}) to sample from a VE model. The results indicate a near halving of sampling time and a remarkable increase in IS within 100 epochs of training, a stark contrast to the performance when VE parameters are traditionally used, as shown by Figure~\ref{fig:Eulers_VE_VP_parameters}. Meanwhile, the Stochastic Sampler emerged as the top performer in terms of IS, as shown in Figure~\ref{fig:Huens_Stochastic}, but with the highest sampling times, as shown in Table~\ref{table:experiment_times}. On the other hand, given that the Heuns' Second Order method with DDIM parameters demonstrates comparable performance with significantly reduced sampling time, the utility of the Stochastic Sampler warrants re-evaluation. Additionally, our investigations into different parameters for the Stochastic Sampler have unveiled a minimal impact of its parameters on performance. Our analysis showed that when the 'ImageNet' parameters in Table~\ref{table:Stochastic_parameters} are used with the Stochastic Sampler on models trained with CIFAR10, the same performance can be obtained with significantly less time, as shown by Tables \ref{table:experiment_times_2},\ref{table:experiment_times} and Figure~\ref{fig:Stochastic_parameters}.

  In fact, the optimal 'CIFAR10' parameters, as recommended by the Karras et al. grid search (\citeyear{karras2022elucidating}),  resulted in increased sampling times without a corresponding increase in performance, suggesting that a meticulous parameter tuning for Stochastic Samplers may not be as critical. The pursuit of optimal sampler performance with the least sampling time leads to the endorsement of Heun's second order solver with DDIM parameters or ODE solvers with VP noise/sampling schedules. On the other hand, the SDE models under study did not fully converge within the training timeframes, hinting at their potential to learn more effectively if granted extended training durations. Evaluating the difference between the samplers at full convergence could be another key to choosing the ideal sampler. Hence, our results show that convergence is contingent on the diffusion process and network design. 

Taken together, our results show that across the DM space, the difference lies in noise/scaling schedules being DDPM/DDIM-based and the diffusion dynamics being DDPM-based. Given the the consistent superior performance of the DDPM-based dynamics over the NCSN-based dynamics in SDE-based models, our results indicate that the diffusion process and the network design factors play the most critical role in enhancing model capabilities.

\section{Discussion}

Our study embarked on a rigorous exploration of DMs through systematic benchmarking across key component choices, specifically targeting the core aspects that underpin their ability to generate high-quality images. We aimed to bring to the surface the intricate dynamics governing DMs and provide a clear path for their future development. Our 

\begin{figure}[H]
    \begin{center}
    \begin{subfigure}{0.5\textwidth}
        \centering
        \includegraphics[width=\textwidth]{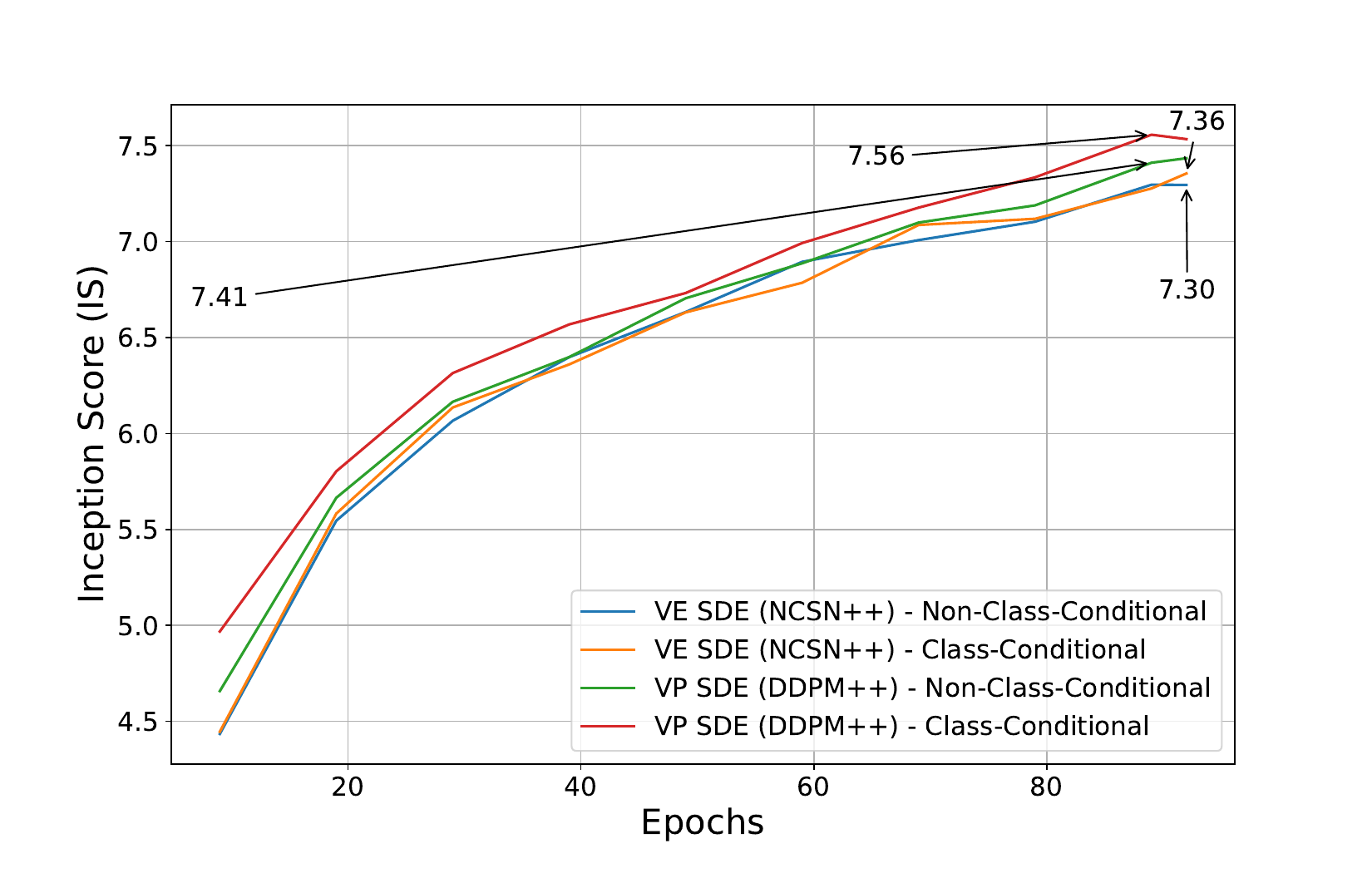}
    \end{subfigure}
    \caption{IS as a function of epochs of VE and VP SDE DMs trained on CIFAR10 with and without class conditioning, and sampled using Heuns 2nd order ODE Solver with DDIM parameters.}
    \label{fig:Huen_DDIM} 
    \end{center}
\end{figure}

collective findings, particularly the consistent superiority of DDPM-based diffusion dynamics, prompt a deeper reflection on the underlying reasons for these outcomes. We offer a renewed perspective on the model properties driving DM progression, highlighting that the next frontier for research in this area lies not in the refinement of ancillary components such as the samplers, but rather in the core design principles of the models themselves like the nature of the diffusion process. Our results highlight that the diffusion process and network design are key to DM performance.
 
The superior performance of using DDPM/DDIM-based diffusion in sampling and DDPM-based diffusion in training is attributed to the refined diffusion dynamics and network design, which appear to be the most consequential in generating high-quality images. These architectural design features were implemented in the transition from NCSNs with DDPMs to VP SDEs \cite{song2020generative, song2020improved, ho2020denoising, song2021scorebased}, enabling the models to be trained for extended periods and to deescalate the number of needed sampling steps. The nuanced differences in performance across guidance, samplers and noise schedules, though informative, pale in comparison to the profound influence of the model's structural design. This divergence from previous studies, where smaller adjustments to guidance, noise schedules or sampler DM configurations were thought to hold more weight, marks a pivotal shift in our understanding of what drives performance. The advancements witnessed in DDPMs with NCSNs, have provided a robust foundation for the subsequent innovations in SDE-based models. This progression is evident from the recurring patterns that use the same fundamental tools observed across different studies. These recurring patterns reflect underlying principles that not only govern the current landscape of DMs but also pave the way for future advancements in the field.

\newpage
\section{Impact Statement}

This paper presents work whose goal is to advance the field of Machine Learning. There are many potential societal consequences of our work, none which we feel must be specifically highlighted here.

\bibliographystyle{icml2024}
\bibliography{main}

\appendix
\renewcommand\thefigure{A.\arabic{figure}}
\renewcommand\thetable{A.\arabic{table}}

\setcounter{figure}{0}

\clearpage

\onecolumn

\section{Appendix}

\subsection{More on the Background of DMs}\label{sec:background_appendix}

\subsubsection{NCSNs}

\paragraph{Diffusion Dynamics}The NCSN framework is designed to enhance score-based generative modeling by incorporating perturbed data using various noise levels and simultaneously estimating scores corresponding to all noise levels through a single conditional score network \cite{song2020generative}. Data is perturbed using a range of noise levels, defined as a positive geometric sequence \(\{\sigma_i\}_{i=1}^L\), where \(\sigma_1 > \sigma_2 > \ldots > \sigma_L\). This perturbation assists in overcoming challenges posed by the manifold hypothesis and low-density regions in the data distribution.

\paragraph{Training} A conditional score network \( s_{\theta}(x, \sigma) \) is trained to estimate the scores of all perturbed data distributions, such that for every \(\sigma \in \{\sigma_i\}_{i=1}^L\), \( s_{\theta}(x, \sigma) \approx \nabla_x \log q_{\sigma}(x) \). The training utilizes denoising score matching, with the objective function for a given \(\sigma\) being:
\[ \mathcal{L}(\theta; \sigma) = \frac{1}{2} \mathbb{E}_{p_{\text{data}}(x)}\mathbb{E}_{x' \sim \mathcal{N}(x, \sigma^2I)}\left[ \left\|s_{\theta}(x', \sigma) + \frac{x' - x}{\sigma^2}\right\|^2 \right]. \]

\paragraph{Sampling} Post-training, annealed Langevin dynamics is utilized to generate samples. The annealed Langevin dynamics algorithm proceeds as follows:

\begin{algorithm}
\caption{Annealed Langevin Dynamics}
\begin{algorithmic}[1]
\REQUIRE{$\{\sigma_i\}_{i=1}^L, \epsilon, T$}
\STATE Initialize $x_0$
\FOR{$i \gets L$ to $1$}
    \STATE $\alpha_i \gets \epsilon \cdot \sigma_i^2 / \sigma_L^2$ $\alpha_i$ is the step size.
    \FOR{$t \gets 1$ to $T$}
        \STATE Draw $z_t \sim \mathcal{N}(0, I)$
        \STATE $x_t \gets x_{t-1} + \frac{\alpha_i}{2} s_{\theta}(x_{t-1}, \sigma_i) + \sqrt{\alpha_i} z_t$
    \ENDFOR
    \STATE $x_0 \gets x_T$
\ENDFOR
\STATE \textbf{return} $x_T$
\end{algorithmic}
\end{algorithm}

This approach ensures effective sampling from distributions with various density regions and multimodal characteristics.

\paragraph{Network Architecture} The network architecture of the NCSN is inspired by successful architectures for dense image prediction tasks such as semantic segmentation. For the purpose of image generation, the NCSN combines the design elements of U-Net with dilated convolutions, and it incorporates instance normalization, particularly a variant tailored to condition on noise levels. The NCSN utilizes conditional instance normalization, denoted as \textit{CondInstanceNorm++}, where the normalization parameters are conditioned on the noise level $\sigma$. This allows the score network $s_{\theta}(x, \sigma)$ to account for $\sigma$ while predicting scores. The normalization for a feature map $x_k$ with mean $\mu_k$ and standard deviation $s_k$ is computed as:
\[
z_k = \gamma[i, k] \frac{x_k - \mu_k}{s_k} + \beta[i, k] + \alpha[i, k] \frac{\mu_k - m}{v},
\]
where $\gamma, \beta \in \mathbb{R}^{L \times C}$ and $\alpha \in \mathbb{R}^{L \times C}$ are learnable parameters, $m$ and $v$ are the mean and standard deviation of $\mu_k$, respectively, and $i$ indexes the noise level $\sigma_i$. Dilated convolutions are employed to expand the receptive field without reducing the resolution of feature maps. This aspect is beneficial for preserving spatial information which is critical in tasks like semantic segmentation and for the generation of detailed images \cite{song2020generative}. The architecture is based on a 4-cascaded RefineNet, which itself is a variant of U-Net comprising ResNet blocks \cite{lin2016refinenet}. The U-Net's characteristic skip connections facilitate the transfer of lower-level features to deeper layers, aiding in the generation of more coherent images. The architecture makes use of pre-activation residual blocks without batch normalization, replaced by \textit{CondInstanceNorm++}. Subsampling is performed only in the first layer, while subsequent downsampling is replaced by dilated convolutions. This choice is made to prevent the loss of spatial resolution in deeper layers. ELU is employed as the activation function throughout the network \cite{clevert2016fast}. The number of filters in the convolutional layers is adapted according to the dataset complexity. For example, in the CelebA and CIFAR-10 datasets, the initial layers have 128 filters, doubling with each cascade, whereas for MNIST, the number of filters is halved to accommodate the lower complexity \cite{song2020generative}.

\subsubsection{NCSNv2}

\paragraph{Diffusion Dynamics Modifications} The original NCSNs utilized a range of noise levels to perturb data, addressing the manifold hypothesis and enabling effective score estimation in low-density data regions. Through the simultaneous estimation of score functions across noise levels via a conditional score network \( s_{\theta}(x, \sigma) \), the framework provided a robust approach for score-based generative modeling. However, the accuracy of the score function estimation was compromised in regions without data, impacting the convergence of the Langevin dynamics. The annealed Langevin dynamics served as a pivotal inference strategy, leveraging the sequential sampling from noise-perturbed distributions to refine the generated samples progressively. Despite its effectiveness, the original NCSN faced challenges with high-resolution image generation, indicating a need for an improved version. NCSNv2 emerged as an advanced iteration of the initial framework, integrating several key techniques to enhance performance and scalability, especially for high-resolution images \cite{song2020improved}. These advancements included:

\begin{itemize}
  \item \textbf{Improved Noise Scale Selection:} A refined technique for the initial noise scale selection was proposed to ensure diversity in the final samples, considering the maximum pairwise distances of training data.
  \item \textbf{Geometric Progression of Noise Levels:} The noise levels were chosen as a geometric progression, facilitating efficient training and better coverage of high-density regions across scales.
  \item \textbf{Optimized Noise Conditioning:} The noise conditioning mechanism was streamlined by parameterizing the NCSN with \( s_{\theta}(x, \sigma) = s_{\theta}(x)/\sigma \), simplifying the model and allowing it to handle a broader range of noise scales.
  \item \textbf{Configured Annealed Langevin Dynamics:} Annealed Langevin dynamics were configured with a theoretically motivated step size schedule, balancing computational efficiency with sample quality.
  \item \textbf{Stabilization with EMA:} Exponential moving average (EMA) was introduced to stabilize the training process, reducing fluctuations in image sample quality and color shifts.
\end{itemize}

\paragraph{Network Architecture Modifications} The NCSNv2 introduces several network architecture improvements over the original NCSN \cite{song2020improved}. Here are the key differences:

\begin{itemize}
  \item \textbf{Number of Layers and Filters}: The NCSNv2 adjusts the number of layers and filters to ensure sufficient capacity and receptive fields for images of different resolutions.

  \item \textbf{ResBlock and RefineBlock}: Instead of using 'CondResBlock' and 'CondRefineBlock' with conditional instance normalization, NCSNv2 uses 'ResBlock' and 'RefineBlock' without any normalization layers. This simplifies the architecture and reduces memory consumption.
  
  \item \textbf{InstanceNorm++}:  InstanceNorm++ is utilized in place of CondInstanceNorm++ which sets the number of classes to 1 and incorporate noise conditioning in the model \cite{song2020improved}.

  \item \textbf{Max Pooling}: NCSNv2 reverts to using max pooling in RefineNet blocks, following the standard architecture more closely \cite{lin2016refinenet, song2020improved}.
  
\end{itemize}

\subsubsection{DDPMs}

\paragraph{Diffusion Dynamics} The diffusion process is a Markov chain that transforms data $x_0$ into noise through a series of latent variables $x_1, \ldots, x_T$:
\begin{align}
q(x_{1:T} | x_0) &= \prod_{t=1}^{T} q(x_t | x_{t-1}), \\
q(x_t | x_{t-1}) &= \mathcal{N}(x_t; \sqrt{1 - \beta_t} x_{t-1}, \beta_t \mathbf{I}),
\end{align}
where $\beta_1, \ldots, \beta_T$ are fixed or learned variance schedules. The forward process is fixed and not learned during training, with the variances $\beta_t$ set as constants. For the reverse process, we consider the untrained time-dependent constants $\sigma^2_t$ as the variance. The mean of the reverse process, $\mu_{\theta}(x_t, t)$, is derived from the forward process posterior mean $\tilde{\mu}_t(x_t, x_0)$.

The architecture of DDPMs is characterized by the reverse process variances and the parameterization of the Gaussian distribution in the reverse process \cite{ho2020denoising}.  To this end, the reverse process mean $\mu_{\theta}(x_t, t)$ is parameterized to predict noise from $x_t$, which simplifies training to a problem akin to denoising score matching. The reverse process, parameterized by $\theta$, is defined similarly as a Markov chain with learned Gaussian transitions:
\begin{align}
p_\theta(x_{0:T}) &= p(x_T) \prod_{t=1}^{T} p_\theta(x_{t-1} | x_t), \\
p_\theta(x_{t-1} | x_t) &= \mathcal{N}(x_{t-1}; \mu_\theta(x_t, t), \Sigma_\theta(x_t, t)).
\end{align}

\paragraph{Training} The model parameters are trained by minimizing the variational bound on the negative log-likelihood, where the forward process introduces noise to the data and the reverse process learns to denoise it:
\begin{equation}\mathbb{E}_{q}[-\log p_\theta(x_0)] \leq \mathcal{L} \end{equation}

\begin{equation}
\mathbb{E}_{q} \left[ -\log p_\theta(x_0| x_T) \right] 
- \sum_{t \geq 1} \mathbb{E}_{q} \left[ \log \frac{p_\theta(x_{t-1}|x_t)}{q(x_t|x_{t-1})} \right] := \mathcal{L} \label{eq:variational_bound}
\end{equation}

where the loss $\mathcal{L}$ is a variational upper bound that decomposes into several terms, including a series of Kullback-Leibler (KL) divergences and a reconstruction loss for the data. Specifically, the loss is composed of:

\begin{itemize}
    \item KL divergences between the forward process posteriors $q(x_{t-1} | x_t, x_0)$ and the reverse process distributions $p_\theta(x_{t-1} | x_t)$, for each time step $t$, which measure how well the reverse process approximates the true posterior distribution of the denoising process.
    \item A reconstruction loss term, often the negative log-likelihood of the data $x_0$ given the first latent variable $x_1$, which encourages the model to accurately reconstruct the data from its noisy representations.
\end{itemize}

\begin{algorithm}
    \caption{Training}
    \begin{algorithmic}[1]
        \REPEAT
        \STATE $x_0 \sim q(x_0)$
        \STATE $t \sim \text{Uniform}(\{1, \ldots, T\})$
        \STATE $\epsilon \sim \mathcal{N}(0, I)$
        \STATE Gradient descent step on $ \nabla_{\theta}|| \epsilon - \epsilon_{\theta}(\sqrt{\overline{\alpha}_t}x_0 + \sqrt{1 - \overline{\alpha}_t}\epsilon, t) ||^2$  where $\overline{\alpha}_t$ is a cumulative product of $1 - \beta_t$
        \UNTIL{converged}
    \end{algorithmic}
\end{algorithm}

A simplified training objective is employed, discarding the weighting in the variational bound and focusing on terms that are more challenging to denoise. The resulting objective emphasizes reconstruction aspects differently compared to the standard variational bound. This reweighting is observed to lead to better sample quality.

\paragraph{Sampling} The Markovian sampling process resembles Langevin dynamics, where the approximator $\epsilon_{\theta}$ serves as a learned gradient of the data density. This process iteratively refines the samples, starting from a standard normal distribution and progressively denoising the data through the learned reverse process.

\begin{algorithm}
    \caption{Markovian Sampler}
    \begin{algorithmic}[1]
        \STATE Initialize $x_T \sim \mathcal{N}(0, I)$
        \FOR{$t = T, T-1, \ldots, 1$}
            \STATE Sample $z \sim \mathcal{N}(0, I)$ if $t > 1$, otherwise set $z = 0$
            \STATE Compute $x_{t-1}$ using the reverse process:
            \begin{equation}
            x_{t-1} = \frac{1}{\sqrt{\alpha_t}} \left( x_t - \frac{1-\alpha_t}{\sqrt{1-\overline{\alpha}_t}} \epsilon_{\theta}(x_t, t) \right) + \sigma_t z
            \end{equation}
        \ENDFOR
        \STATE \textbf{return} $x_0$
    \end{algorithmic}
\end{algorithm}

The parameter $\alpha_t$ is a time-dependent factor that adjusts the noise level. The function $\epsilon_{\theta}$ represents the output of the neural network given the noised data $x_t$ and time step $t$. The sampler's goal is to produce high-fidelity samples that are representative of the original data distribution.

\paragraph{Network Architecture}The network design for the DDPMs is inspired by a U-Net architecture \cite{ronneberger2015unet}, incorporating elements from the PixelCNN++ design \cite{salimans2017pixelcnn}. The network employs a series of convolutional layers structured in a manner that allows for both down-sampling and up-sampling paths, enabling the capture of features at multiple scales. Group normalization \cite{Wu_2018_ECCV} is used throughout the network to stabilize the training process by normalizing the inputs across grouped subsets of channels. This choice replaces the commonly used batch normalization due to its benefits in small batch size scenarios. Self-attention mechanisms \cite{vaswani2023attention} are introduced at the 16 $\times$ 16 feature map resolution, facilitating the model's capacity to capture long-range dependencies within the data. The network is also conditioned on the diffusion time variable $t$, using sinusoidal position embeddings \cite{vaswani2023attention}, to guide the generation process through different stages of the diffusion. The architecture's flexibility allows for the use of residual connections, which help in mitigating the vanishing gradient problem by allowing gradients to flow through alternate pathways \cite{ho2020denoising}.

\subsubsection{DDIM}

DDIMs accelerates the sampling process of DDPMs while preserving sample quality. DDIMs are conceptualized as a non-Markovian generalization of DDPMs, where the forward process does not strictly follow a Markov chain but rather a sequence that allows for more efficient reverse mapping from noise to data \cite{song2022denoising}.

DDIMs are trained with the same objective as DDPMs, making use of a surrogate objective function corresponding to the variational lower bound. This is expressed as:

\begin{equation}
\max_{\theta} \mathbb{E}_{q(x_0)} [\log p_{\theta}(x_0)] \leq 
    \max_{\theta} \mathbb{E}_{q(x_0, x_1, \ldots, x_T)} [\log p_{\theta}(x_0:T) - \log q(x_1:T | x_0)], 
\end{equation}
where $p_{\theta}(x_0:T)$ denotes the joint distribution over the data and latent variables, and $q(x_1:T | x_0)$ is the fixed inference process.

The reverse DDIM process is constructed to denoise the data, progressively reducing noise to obtain a clean sample from an initial noise distribution $p_{\theta}(x_T)$. The reverse process is modeled with Gaussian conditionals, as shown in the equation:
\begin{equation}
    p_{\theta}(x_0:T) = p_{\theta}(x_T) \prod_{t=1}^{T} p^{(t)}_{\theta}(x_{t-1}|x_t),
\end{equation}
where each $p^{(t)}_{\theta}(x_{t-1}|x_t)$ is a Gaussian distribution that depends on the parameters $\theta$ learned during training.

The generative process of DDIMs can be related to the Euler method for solving ordinary differential equations (ODEs), drawing a parallel with Neural ODEs \cite{song2022denoising}. This relationship is formalized in the proposed ordinary differential equation:
\begin{equation}
    dx(t) = \epsilon_{\theta} \left( \frac{x(t)}{\sqrt{\sigma^2 + 1}} \right) d\sigma(t),
\end{equation}
where $\epsilon_{\theta}$ is the model attempting to predict the noise, and $\sigma(t)$ is a continuous, increasing function starting from zero.

\subsubsection{Linear Noise Schedule}

Ho et al. (\citeyear{karras2022elucidating}) sets the variance schedule to increase linearly from $\beta_1 = 10^{-4}$ to $\beta_T = 0.02$ for $T = 1000$ . We experimented with different linear noise schedules by adding a scaling factor that changes $\beta_1$ and $\beta_T$. By using scaling factors of 0.25, 0.5, 0.75 and 1 we experimented with the following combinations of ($\beta_1$, $\beta_T$) : [($2.5\times10^{-5},0.005 ),(5\times10^{-5}, 0.001),(7.5\times10^{-5}, 0.015),(10^{-4}, 0.02)$] respectively.

\subsubsection{Cosine Noise Schedule}

The cosine noise schedule is defined by a smooth, non-linear progression of the variance terms \(\{\beta_t\}\), designed to maintain a more consistent information flow throughout the diffusion process \cite{nichol2021improved}. It is parameterized as follows:
\begin{equation}
\hat{\alpha}_t = \frac{f(t)}{f(0)}, \quad f(t) = \cos\left(\frac{t/T + s}{1 + s} \cdot \frac{\pi}{2}\right)^2 
\end{equation}
\begin{equation}
\beta_t = 1 - \frac{\alpha_t}{\alpha_{t-1}}
\end{equation}
where \( t \) indexes the diffusion steps from 1 to \( T \), \( s \) is a small offset to prevent vanishing noise at the initial steps that is set to 0.008 such that $sqrt(\beta_0)$ is slightly smaller than the pixel
bin size, and \( T \) is the total number of diffusion steps. The schedule starts with a gentle introduction of noise, maintains a nearly linear increase in the middle phase, and tapers off towards the end. This gradual change prevents the abrupt destruction of information and allows for better retention of detail in generated samples. The noise schedule is applied inversely during the reverse diffusion process to recover the denoised sample from the latent noise \cite{nichol2021improved}.

\subsubsection{Sigmoid Noise Schedule}

 The sigmoid noise schedule is characterized by parameters \( \eta \), is denoted as \( \gamma_{\eta}(t) \) and is trained to output a value that directly corresponds to the variance of the noise at time \( t \) \cite{jabri2023scalable, chen2023analog, kingma2023variational, chen2023importance}. The variance at each time step is thus defined as:

\begin{equation}
\sigma_t^2 = \text{sigmoid}(\gamma_{\eta}(t)) \quad (3)
\end{equation}

The monotonic nature of \( \gamma_{\eta}(t) \) ensures that as the diffusion process progresses, the noise schedule is smoothly adapted based on the learned parameters, leading to a more flexible and potentially more efficient diffusion process. To maintain the variance-preserving property in both discrete-time and continuous-time DMs, the following transformation is implemented:

\begin{equation}
\alpha_t = \sqrt{1 - \sigma_t^2} \quad (4)
\end{equation}

which allows for a consistent signal-to-noise ratio (SNR) to be implemented throughout the diffusion process. The SNR at time \( t \) and the squared value of \( \alpha_t \) are both functions of \( \gamma_{\eta}(t) \), simplifying the relationship to:

\begin{equation}
\alpha_t^2 = \text{sigmoid}(-\gamma_{\eta}(t)) \quad (5)
\end{equation}

\begin{equation}
\text{SNR}(t) = \exp(-\gamma_{\eta}(t)) \quad (6)
\end{equation}

By learning the noise schedule through \( \gamma_{\eta}(t) \),  the diffusion process is dynamically adapt, potentially improving the quality of generated samples and offering a more nuanced control over the diffusion trajectory \cite{jabri2023scalable}.

\subsubsection{SDE-Based Diffusion}

\paragraph{Diffusion Dynamics} SDEs provide a framework for modeling the continuous transformation of data distributions. The forward SDE starts from a data distribution and adds noise over time \cite{song2021scorebased}, transforming it into a simple distribution, typically a Gaussian:

\begin{equation}
    d\mathbf{x} = \mathbf{f}(\mathbf{x}, t)dt + \mathbf{G}(t)d\mathbf{w},
\end{equation}

where $\mathbf{x} \in \mathbb{R}^d$ is the state, $\mathbf{f}: \mathbb{R}^d \times [0,T] \rightarrow \mathbb{R}^d$ is the drift coefficient, $\mathbf{G}: [0,T] \rightarrow \mathbb{R}^{d \times d}$ is the diffusion coefficient, $d\mathbf{w}$ is the Wiener process, and $T$ is the terminal time.

The reverse-time SDE recovers the data from the noise by reversing this process, which requires knowledge of the score of the data distribution at each point in time:

\begin{equation}
    d\mathbf{x} = [\mathbf{f}(\mathbf{x}, t) - \mathbf{G}(t)\mathbf{G}(t)^\top \nabla_{\mathbf{x}} \log p_t(\mathbf{x})]dt + \mathbf{G}(t)d\bar{\mathbf{w}},
\end{equation}

where $p_t(\mathbf{x})$ is the probability density function of $\mathbf{x}$ at time $t$, and $d\bar{\mathbf{w}}$ is the reverse-time Wiener process.

\paragraph{Training} To train SDE-based models, the score function is estimated, $\nabla_{\mathbf{x}} \log p_t(\mathbf{x})$, using score matching with a neural network:

\begin{equation}
    \theta^* = \arg \min_{\theta} \mathbb{E}_{t \sim \mathcal{U}(0,T)} \mathbb{E}_{\mathbf{x}(0) \sim p_0} \mathbb{E}_{\mathbf{x}(t) | \mathbf{x}(0)} \\ \left[ \lambda(t) \left\| s_{\theta}(\mathbf{x}(t), t) - \nabla_{\mathbf{x}(t)} \log p_t(\mathbf{x}(t) | \mathbf{x}(0)) \right\|^2 \right]
\end{equation}

where $s_{\theta}(\mathbf{x}, t)$ is the score function approximated by the neural network with parameters $\theta$, and $\lambda(t)$ is a positive weighting function.

\paragraph{Sampling} Given a trained score-based model, sampling involves numerically integrating the reverse-time SDE. Various samplers can be employed, such as Euler-Maruyama discretization for the SDE or predictor-corrector sampler, which refine the numerical SDE integration with Markov Chain Monte Carlo (MCMC) steps guided by the score function.

\paragraph{Network Architecture} The network design of SDE-based DMs incorporated several enhancements \cite{song2021scorebased, karras2022elucidating}:

\begin{itemize}
    \item \textbf{Anti-Aliasing in Upsampling and Downsampling:} Images were upsampled and downsampled using Finite Impulse Response (FIR) filtering, following the implementation in StyleGAN-2.
    \item \textbf{Rescaling Skip Connections:} All skip connections were rescaled by a factor of $\frac{1}{\sqrt{2}}$, a technique proven effective in advanced GAN models.
    \item \textbf{Residual Blocks Replacement:} The original residual blocks in the DDPM model were replaced with those from BigGAN, enhancing model expressiveness.
    \item \textbf{Increased Residual Blocks:} The number of residual blocks per resolution was increased from 2 to 4, augmenting the model's capacity.
    \item \textbf{Progressive Growing Architecture:} The architecture incorporated progressive growing techniques for both input and output layers, defined in the style of StyleGAN-2.
    \item \textbf{Attention Mechanisms:} Attention layers with multiple heads were integrated at different resolutions to enhance feature capture capabilities.
    
\end{itemize}

\subsubsection{ODE Solvers}

\paragraph{Sampling with ODE Solvers} Sampling in stochastic DMs involves solving an SDE that describes the evolution of data through noise levels. For practical implementation, a deterministic counterpart, known as an ODE, is often solved numerically. Euler's method, a first-order method \cite{kloeden2013numerical}, is given by the update rule:
\begin{equation}
    x_{i+1} = x_i + f(x_i, t_i)\Delta t,
\end{equation}
where \( f(x_i, t_i) \) represents the ODE function evaluated at step \( i \), and \( \Delta t \) is the step size.

Heun's method, which offers a better tradeoff between accuracy and computational cost \cite{ascher1998computer}, incorporates an additional corrector step to account for changes in the derivative:
\begin{align}
    k_1 &= f(x_i, t_i), \\
    k_2 &= f(x_i + k_1 \Delta t, t_i + \Delta t), \\
    x_{i+1} &= x_i + \frac{1}{2}(k_1 + k_2)\Delta t.
\end{align}

The deterministic sampling process in the Karras et al. utilizes Heun's second-order method, which is an improved form of the Euler method (\citeyear{karras2022elucidating}). The process starts with the generation of an initial noise image $x_0$ sampled from a normal distribution with a variance determined by the maximum noise level, $\sigma_{\text{max}}^2$. The noise reduction process involves multiple iterations, where each iteration denoises the image to a lower noise level, following a predefined schedule $\sigma(t)$. In each iteration $i$, the method calculates the derivative $d_i$ of the image $x_i$ with respect to the current noise level $\sigma(t_i)$. This derivative incorporates changes in both the noise level and the scaling factor of the image. The Euler step is then applied to progress from the current noise level $\sigma(t_i)$ to the next level $\sigma(t_{i+1})$. If the subsequent noise level $\sigma(t_{i+1})$ is non-zero, a second-order correction is applied using Heun's method to refine the estimate of $x_{i+1}$. The process continues iteratively, reducing the noise level at each step, until the final noise-free image $x_N$ is obtained. The choice of noise levels and the step sizes are crucial for the performance of the sampler. In our implementation, these are carefully chosen to minimize truncation errors while ensuring efficient computation.

\paragraph{Stochastic Sampler} The Stochastic Sampler extends the deterministic Heun's method by incorporating stochastic elements to improve sample diversity and quality. The Stochastic Sampler integrates a Langevin-like process for noise injection and removal \cite{karras2022elucidating}. The key feature of this Stochastic Sampler is the alternating use of noise addition and removal steps. This approach helps in maintaining the desired noise distribution at each step of the sampling process. In each iteration $i$, the sampler first increases the noise level of the current sample $x_i$ by a factor $\gamma_i$, creating an intermediate noise-augmented sample $\hat{x}_i$. This increase in noise level is controlled and bounded to avoid excessive noise addition. Following the noise augmentation, the method performs a backward ODE step from the increased noise level $\hat{\sigma}(t_i)$ to the next lower level $\sigma(t_{i+1})$. This step involves evaluating the derivative of $\hat{x}_i$ at $\hat{\sigma}(t_i)$ and then applying Heun's method to estimate the sample at the next iteration, $x_{i+1}$.

\subsubsection{ODE Solvers Parameters}

\paragraph{ODE Solvers} The ODE integration requires a discretization scheme to define the time steps for the numerical solution. The time steps \( \{t_0, t_1, \ldots, t_N\} \), where \( t_N = 0 \), are chosen to ensure that the noise levels decrease monotonically. A parameterized scheme is employed where the time steps correspond to a sequence of noise levels \( \{\sigma_0, \sigma_1, \ldots, \sigma_N\} \), ensuring that truncation errors are minimized at each step. The choice of \( \Delta t \) and the scheduling of \( \sigma(t) \) and \( s(t) \) are crucial for the quality and efficiency of the generated samples. The methods above demonstrate that with careful design of the ODE solver and its parameters, SOTA sampling efficiency can be achieved in DMs \cite{karras2022elucidating}.

\begin{table}[htbp]
\begin{center}
\centering
\vspace{0.1in}
\caption{Parameters of SDE samplers (ODE solvers) \cite{karras2022elucidating}.}
\label{tab:sampling_methods}
\vspace{0.1in}
\begin{tabular}{cccc}
\hline
 & \textbf{VP} & \textbf{VE} & \textbf{Huen solver with DDIM parameters} \\
\hline
ODE solver & Euler & Euler & 2nd order Heun \\
Time steps & $t_i < N$ & $t$ & $(\sigma_{\text{max}}^{\rho} + \frac{i}{N-1}(\sigma_{\text{min}}^{\rho} - \sigma_{\text{max}}^{\rho}))^{\rho}$ \\
 & $1 + \frac{i}{N-1}(\epsilon_s - 1)$ & & \\
Noise Schedule & $\sigma(t)$ & $\sqrt{t}$ & $t$ \\
 & $\sqrt{\frac{1}{e^{\frac{1}{2}\beta_d t^2 + \beta_{\text{min}} t} - 1}}$ & & \\
Scaling Schedule & $s(t)$ & $1$ & $1$ \\
 & $1 / \sqrt{e^{\frac{1}{2}\beta_d t^2 + \beta_{\text{min}} t}}$ & & \\
\hline
\end{tabular}
\label{table:Sampling_parameters}
\end{center}
\end{table}

\paragraph{Stochastic Sampler} The stochasticity in the Stochastic Sampler is introduced through the controlled addition of noise and is crucial for correcting any errors accumulated in earlier sampling steps. However, excessive stochasticity can lead to loss of detail and color oversaturation in the generated images. To mitigate this, the stochastic elements are confined within a specific range of noise levels and employ heuristics for noise addition. The Stochastic Sampler utilizes several key parameters to control the noise addition and reduction process. These parameters are crucial for balancing the stochastic elements in the sampling process, ensuring the generation of high-quality samples. The primary parameters include:

\begin{itemize}
    \item \textbf{$S_{churn}$}: This parameter controls the overall amount of stochasticity introduced in the sampling process. It dictates the extent to which new noise is added to the image in each iteration. A higher value of \texttt{$S_{churn}$} increases the stochasticity, which can enhance diversity but may also lead to loss of detail in the generated images.

    \item \textbf{$S_{tmin}$} and \textbf{$S_{tmax}$}: These parameters define the range of noise levels within which stochasticity is enabled. Stochastic operations are performed only when the current noise level $\sigma(t_i)$ lies within this range. By confining stochasticity to a specific noise level range, undesirable effects can be prevented like color oversaturation at very low or high noise levels.

    \item \textbf{$S_{noise}$}: The \texttt{$S_{noise}$} parameter slightly inflates the standard deviation for the newly added noise. This adjustment helps counteract the potential loss of detail caused by a tendency of the denoiser to remove slightly too much noise. A value slightly above 1 for \texttt{$S_{noise}$} has been found effective in maintaining image quality.

\end{itemize}

\begin{table}[htbp]
\begin{center}
\centering
\caption{Grid searched Stochastic Sampler parameters for CIFAR10 and ImageNet by Karras et al. (\citeyear{karras2022elucidating}).\\}
\label{tab:stochastic_sampling_params}
\vspace{0.1in}
\begin{tabular}{ccccc}
\hline
\textbf{Parameter} & \textbf{CIFAR-10} & \textbf{CIFAR-10} & \textbf{ImageNet}\\
 & \textbf{VP} & \textbf{VE} & \textbf{VP/VE}\\
\hline
\( S_{\text{churn}} \) & 30 & 80 & 40\\
\( S_{\text{min}} \) & 0.01 & 0.05 & 0.05\\
\( S_{\text{max}} \) & 1 & 1 & 50 \\
\( S_{\text{noise}} \) & 1.007 & 1.007 & 1.003\\
\hline
\end{tabular}
\label{table:Stochastic_parameters}
\end{center}
\end{table}

These parameters collectively ensure that the stochastic sampler effectively balances the introduction of randomness with the maintenance of image quality and fidelity to the target data distribution \cite{karras2022elucidating}.

\subsection{Training Parameters}

Our research involved training various DMs, namely NCSNv2, DDPM, NCSN++, and DDPM++, on multiple datasets including CIFAR10 and FFHQ. This section details the specific hyperparameters and configurations employed in these training processes.

\begin{table}[h]
\begin{center}
\centering
\caption{Hyperparameters for NCSNv2, DDPM, NCSN++, and DDPM++ models trained on CIFAR10, FFHQ, and ImageNet datasets.}
\vspace{0.1in}
\scalebox{0.9}{\begin{tabular}{|c|c|c|c|c|}
\hline
\textbf{Parameter} & \textbf{NCSNv2} & \textbf{DDPM} & \textbf{NCSN++} & \textbf{DDPM++} \\
\hline
Batch Size & 32 & 32 & 32 & 32 \\
EMA Beta & 0.999 & 0.999 & 0.999 & 0.999 \\
Dropout Probability & 13 \% & 13 \% & 13 \% & 13 \%\\
Learning Rate & $1 \times 10^{-4}$ & $10 \times 10^{-4}$ & $10 \times 10^{-4}$ & $10 \times 10^{-4}$ \\
Optimizer & ADAM & ADAM & ADAM & ADAM \\
DDPM Timesteps & - & 1000 & - & - \\
Markovian Sampling Steps & - & 1000 & - & - \\
DDIM Sampling Steps & - & 999 & - & - \\
Depth & - & \{4 - For All Datasets\} & - & - \\
Attention Resolutions & - & \{Dataset Specific\} & \{16\} & \{16\} \\
GPUs & 1 $\times$ A100 & 1 $\times$ A100 & 1 $\times$ A100 & 1 $\times$ A100 \\
Resampling Filter & Bilinear & Box & Bilinear & Box \\
Noise Embedding & - & Positional & Fourier & Positional \\
Skip Connections in Encoder & Residual & - & Residual & - \\

\hline
\end{tabular}}
\label{tab:hyperparams}
\end{center}
\end{table}

Table~\ref{tab:hyperparams} outlines the hyperparameters used for each model. A consistent batch size of 32 was maintained across all models and datasets. The EMA Beta was set to 0.999 for all models. To mitigate overfitting, a dropout probability of 13\% was applied. The learning rate was set at $1 \times 10^{-4}$ for NCSNv2 and $10 \times 10^{-4}$ for DDPM, NCSN++, and DDPM++. All models employed the ADAM optimizer. Specific to the DDPM model, we used 1000 timesteps for Markovian Sampling, and 999 steps for DDIM Sampling. The depth parameter was set to 4 for all datasets in the DDPM model. Attention resolutions were dataset-specific for DDPM and set to 16 for NCSN++ and DDPM++. Each model was trained on a single NVIDIA A100 GPU. Resampling filters varied between bilinear and box types, and noise embedding techniques included positional and Fourier embeddings. Skip connections in the encoder were residual for NCSNv2 and NCSN++.
\vspace{-28pt}

\begin{table}[h]
\begin{center}
\centering
\vspace{0.1in}
\caption{Configuration parameters for the pretrained classifier used for Classifier Guidance in DDPMs with CIFAR10.}
\vspace{0.1in}
\begin{tabular}{|c|c|}
\hline
 & \textbf{CIFAR10} \\
\hline
Image Size & 32 \\
\hline
Channel Multiplier & (1, 2, 4) \\
\hline
Attention Resolutions (pixels) & 16, 8, 4 \\
\hline
Model Channels & 64 \\
\hline
Number of Res Blocks & 2 \\
\hline
Num Head Channels & 64 \\
\hline
Scale-Shift Norm & True \\
\hline
Resblock Up/Down & True \\
\hline
Batch Size (Training Arg) & 32 \\
\hline
Number of Epochs (Training Arg) & 200 \\
\hline
Classifier Scale (Training Arg) & 1 \\
\hline
GPUs & 1 $\times$ A100 \\ 
\hline
\end{tabular}
\label{tab:classifier_params}
\end{center}
\end{table}
\vspace{-10pt}

In addition to the primary model parameters, we also configured a pretrained classifier for Classifier Guidance in DDPMs, specifically for the CIFAR10 dataset.  The architecture of the classifier is based on the downsampling trunk of the U-Net model, with attention pooling introduced at the $8 \times 8$ layer to generate the final output \cite{dhariwal2021diffusion}. The classifier was trained on noisy CIFAR10 images. As detailed in Table~\ref{tab:classifier_params}, the classifier configuration included an image size of 32, a channel multiplier of (1, 2, 4), attention resolutions at 16, 8, and 4 pixels, and model channels set to 64. The classifier comprised 2 residual blocks, with 64 head channels. Scale-shift normalization was enabled, along with resblock up/down configurations. For training, the batch size was set to 32 with a total of 200 epochs, and a classifier scale of 1 was used. Similar to the DMs, the classifier was trained on a single NVIDIA A100 GPU. These detailed configurations and parameters were crucial in ensuring the robustness and effectiveness of our DMs across various datasets and in achieving the desired comparability in our experiments.

\subsection{Compute Comparison}

Our extensive experiments demonstrate the efficiency of various sampling methods when applied to DMs trained on the CIFAR10 dataset. We ensured optimal resource utilization by achieving 100\% GPU utilization across all experiments.

\vspace{-10pt}
\begin{table}[H]
\begin{center}
\centering
\caption{\label{table:experiment_times}  Training and Sampling times of DM variants trained on CIFAR10}
\scalebox{0.7}{\begin{tabular}{|p{1.2in}||p{1.2in}|p{1.2in}|p{1.2in}|p{1.2in}|p{1.2in}|
p{1.2in}|}
 \hline\hline
\bf DDPM Variant & \bf Sigmoid Schedule & \bf Cosine Schedule & \bf Linear Schedule Scaling Factor: 0.25  & \bf Linear Schedule Scaling Factor: 0.5  & \bf Linear Schedule Scaling Factor: 0.75 & \bf Linear Schedule Scaling Factor: 1\\ [0.5ex] 
 \hline
 Training Time (90 epochs) & 298 min &  300 min & 298 min & 329 min & 299 min & 298 min \\ 
  \hline\hline
 \bf DDPM Sampler & \bf Markovian Sampler  & \bf DDIM Sampler & \bf Markovian Sampler + Classifier Guidance & \bf DDIM Sampler  + Classifier Guidance & - & - \\ [0.5ex] 
 \hline
Sampling Time (10k images) & 90 min & 89 min & 132 min & 231 min & - & - \\ 
 \hline\hline
\bf SDE Variant & \bf VP (DDPM++) Class-Conditional & \bf VP (DDPM++) Class Non-Conditional & \bf VE (NCSN++) Class-Conditional  & \bf VE (NCSN++) Class Non-Conditional & - & -\\ [0.5ex] 
 \hline
 Training Time (90 epochs) & 400 min &  400 min & 408 min & 408 min & - & - \\ 
 \hline\hline
 \bf SDE Sampler &  \bf Euler's Method &  \bf  Reimplemented Euler's method &  \bf Heuns Second Order Method VP/VE &  \bf Heuns Second Order Method with DDIM parameters &  \bf Stochastic Class Conditional & \bf Stochastic Class Non-Conditional VP/VE\\ [0.5ex] 
 \hline
 Sampling Time (10k images) & 7 min & 79 min & 39 min / 133 min & 7 min & 171 min & 82 min / 400 min \\ 
 \hline
\end{tabular}
}
\end{center}
\end{table}

As shown in Table~\ref{table:experiment_times}, different noise scheduling techniques were employed to analyze their impact on the training time of DM variants. The results indicate minimal variation in training times across different schedules in DDPMs, with times ranging from 298 minutes to 329 minutes for 90 epochs, demonstrating a relatively stable training duration regardless of the noise schedule employed. In terms of sampling efficiency, our results suggest that the implementation of Classifier Guidance leads to a noticeable increase in sampling time.

Comparing the SDE variants, the VP (DDPM++) model and the VE (NCSN++) model displayed similar training times of 400 and 408 minutes, respectively, for 90 epochs. This slight increase in training time for the VE model could be attributed to the different noise scheduling approach inherent to the NCSN++ architecture. Evaluating the efficiency of SDE samplers, in addition to being among the best performers between the ODE Solvers in terms of IS, the Heun's second-order method with DDIM parameters was found to be exceptionally time-efficient, taking only 7 minutes to sample 10k images. 

\begin{table}[H]
\begin{center}
\centering
\vspace{0.1in}
\caption{\label{table:experiment_times_2} Sampling times when various parameters were implemented to sample from on VP/VE SDE DMs trained on CIFAR10}
\vspace{0.1in}
\begin{tabular}{|p{1.2in}||p{1.2in}|p{1.2in}|p{1.2in}|p{1.2in}|p{1.2in}|
p{1.2in}|}
 \hline\hline
 \bf SDE Sampler &  \bf Reimplemented Euler's method with VP noise/scale schedules &  \bf Heuns second Order method with VP noise/scale schedules & \bf Stochastic sampler with ImageNet Parameters\\ [0.5ex] 
 \hline
 Sampling Time (10k images) & 79 min & 39 min & 74 min \\ 
 \hline
 
\end{tabular}
\end{center}
\end{table}

Table~\ref{table:experiment_times_2} presents a comparison of sampling times utilizing different ODE solvers with VP/VE SDE DMs. Notably, the Heun's second-order method with VP noise/scale schedules outperformed the other samplers, requiring only 39 minutes to sample 10k images when used with both VE and VP models, which is approximately half the time taken by the re-implemented Euler's solver with VP noise/scale schedules (79 min) and significantly faster than the stochastic sampler with ImageNet parameters (74 min).

\subsection{Effect of Number of Generated Samples on IS}

\begin{figure}[h]
\begin{adjustwidth}{-2cm}{-2cm}
    \centering
    \begin{center}
    \begin{subfigure}{0.5\textwidth}
        \centering
        \includegraphics[width=\textwidth]{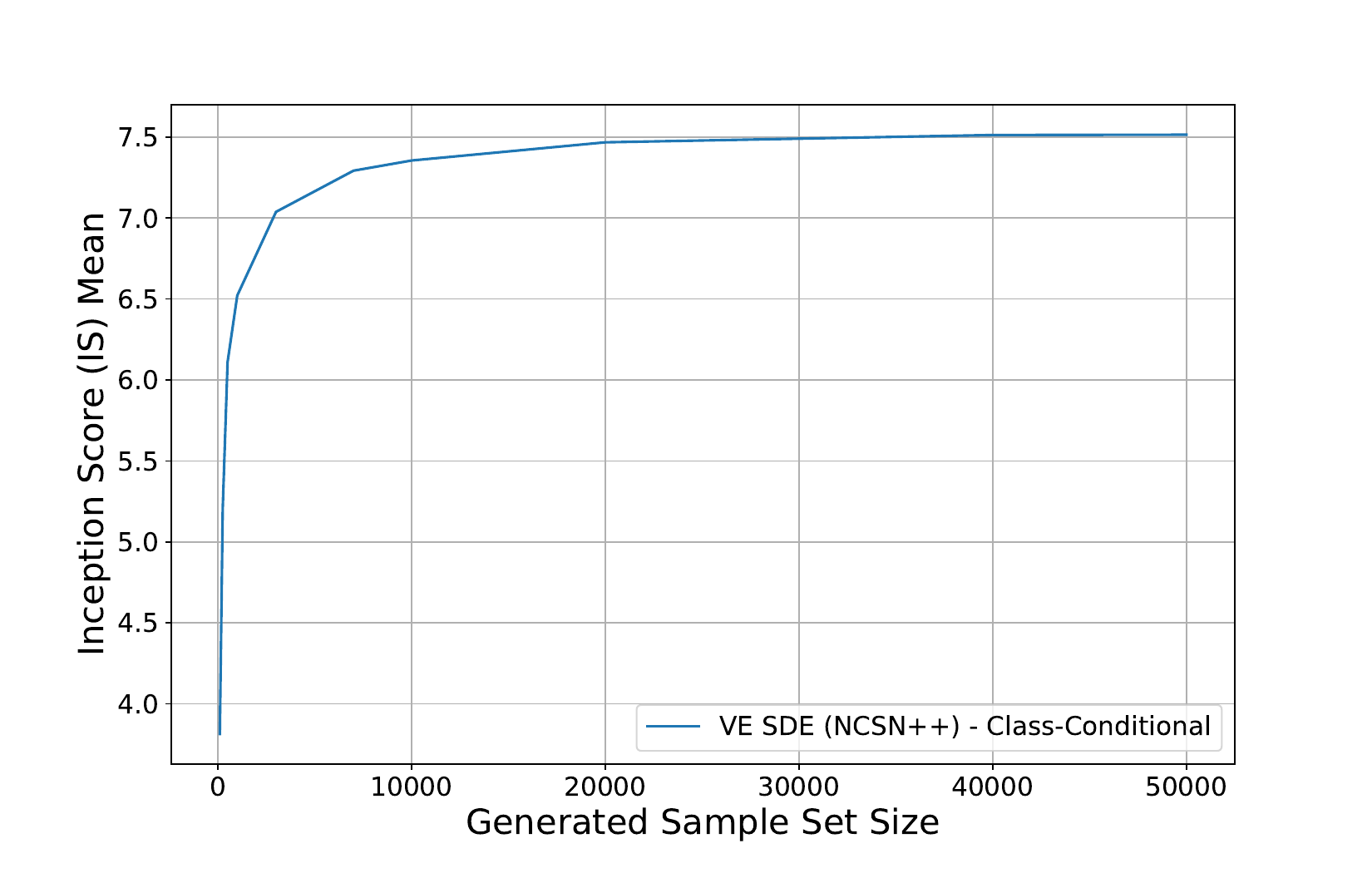}
        \caption{}
    \end{subfigure}
        \begin{subfigure}{0.5\textwidth}
        \centering
        \includegraphics[width=\textwidth]{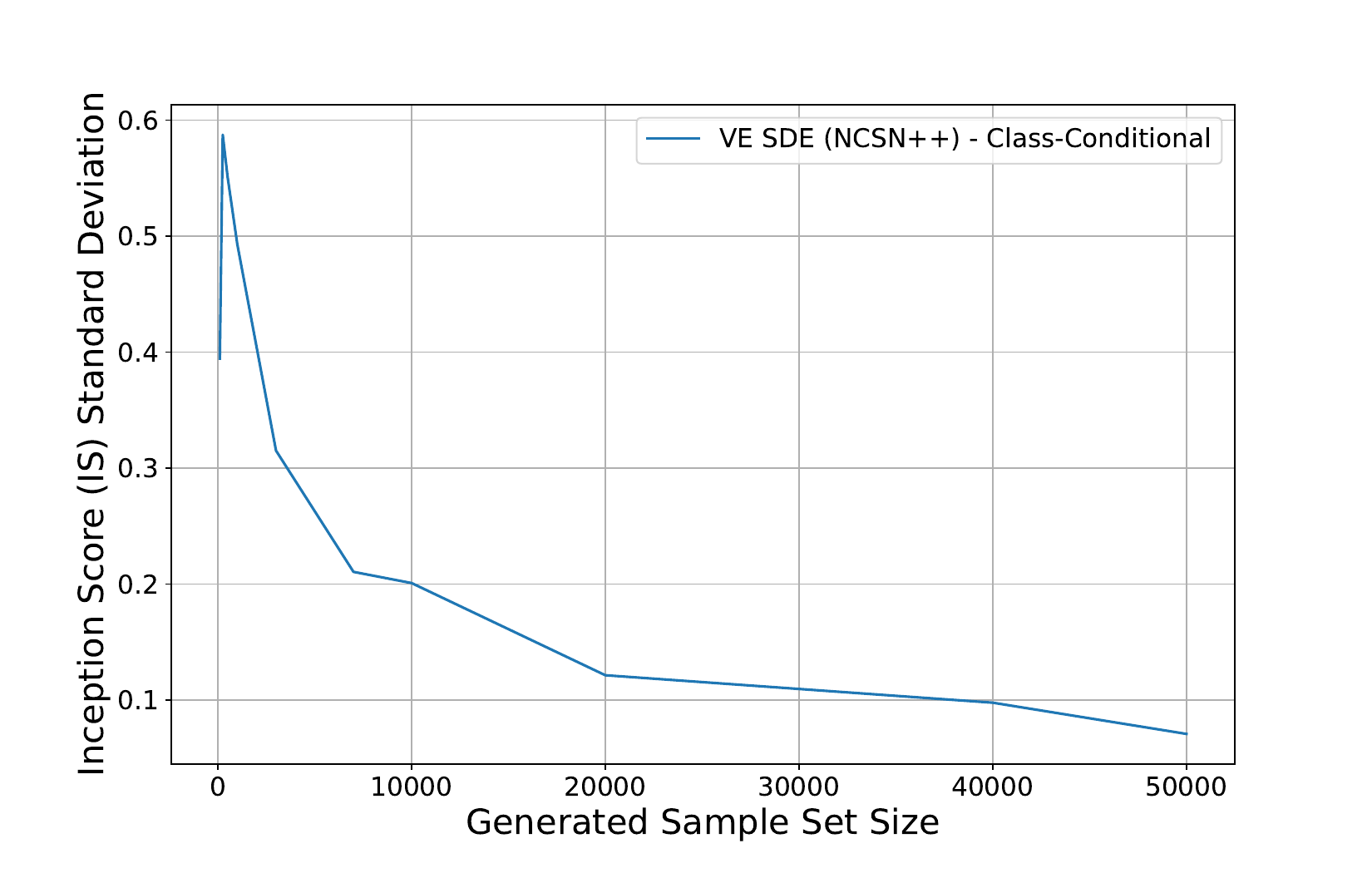}
        \caption{}
    \end{subfigure}
    \end{center}
    \end{adjustwidth}
    \caption{a) IS mean and b) IS standard deviation  as a function of number of samples of Non-Class Conditional VE SDE model trained on CIFAR10 and sampled with the Karras et al. reimplemented Euler's ODE solver (\citeyear{karras2022elucidating}).}
    \label{fig:IS_mean_std}
\end{figure}

As shown by Figure~\ref{fig:IS_mean_std}, the IS mean increases with the number of generated samples. The IS mean starts from a lower value and exhibits a steep rise as the number of generated samples increases until it reaches a plateau. This trend suggests that after a certain point, generating more samples does not significantly increase the mean IS, indicating that the model's generative quality becomes consistent after a threshold number of samples. From the plot, it is apparent that the IS mean stabilizes after approximately 10,000 generated samples, which aligns with the notion that a generated sample size of 10,000 is sufficient for reliable IS evaluation.

The standard deviation of the IS decreases as the number of generated samples increases, as shown by Figure~\ref{fig:IS_mean_std}. There is a sharp decline initially, indicating a rapid increase in consistency of the model's generative performance. As the number of samples increases, the standard deviation flattens out, suggesting minimal variance and hence greater reliability in the IS.
Consistent with the IS mean plot, the standard deviation of IS also stabilizes after reaching the threshold of 10,000 generated samples, reinforcing the conclusion that beyond this point, additional samples do not substantially contribute to the measure of generative quality.

\subsection{Markovian Sampling under Misguided Diffusion}

\vspace{-24pt}

\begin{figure}[H] 
    \begin{adjustwidth}{-2cm}{-2cm}
    \begin{center}
    \begin{subfigure}{0.5\textwidth}
        \centering
        \includegraphics[width=\textwidth]{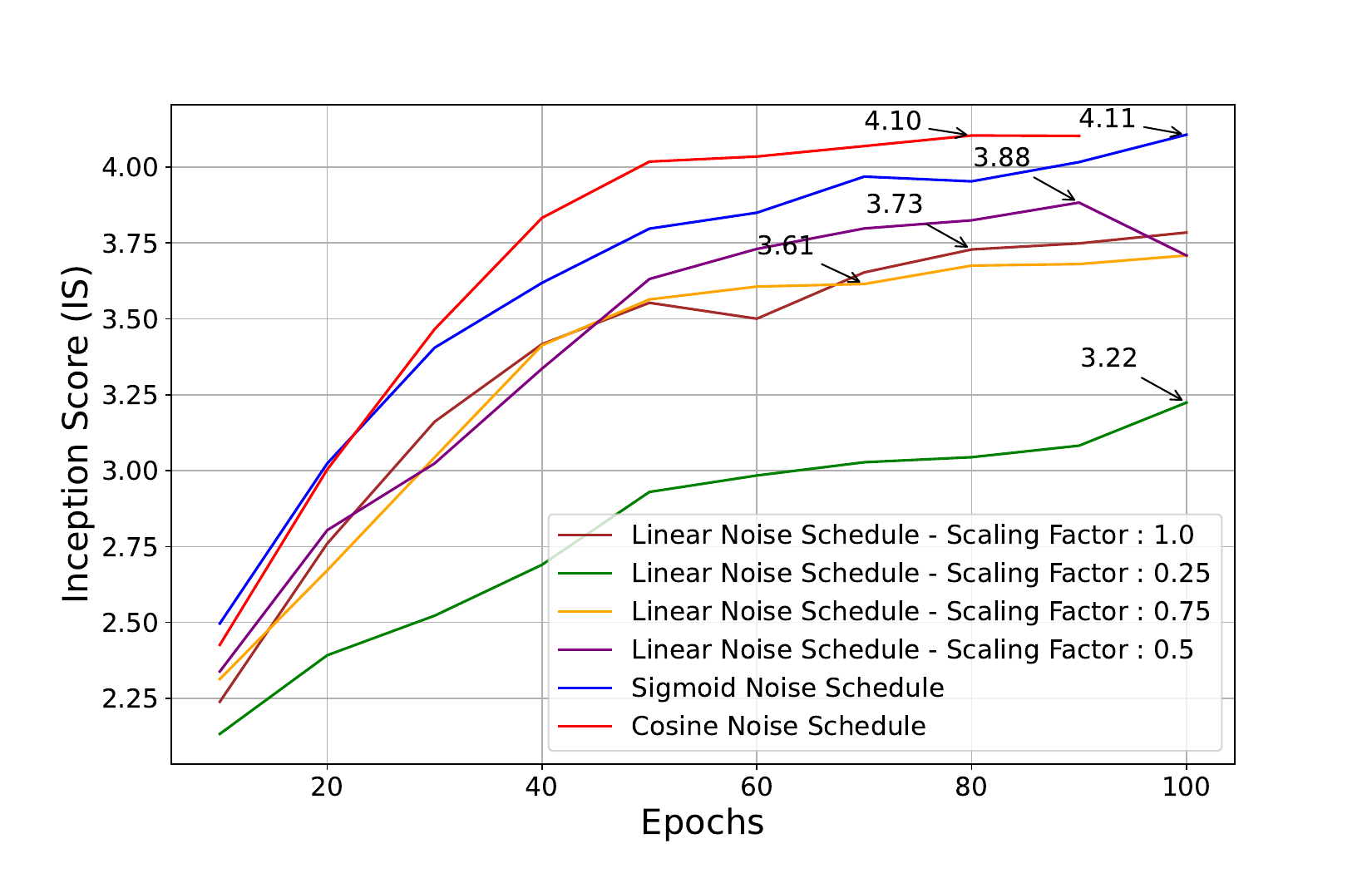}
        \subcaption{}
        \label{fig:Guidance_2}
    \end{subfigure}
    \begin{subfigure}{0.5\textwidth}
        \centering
        \includegraphics[width=\textwidth]{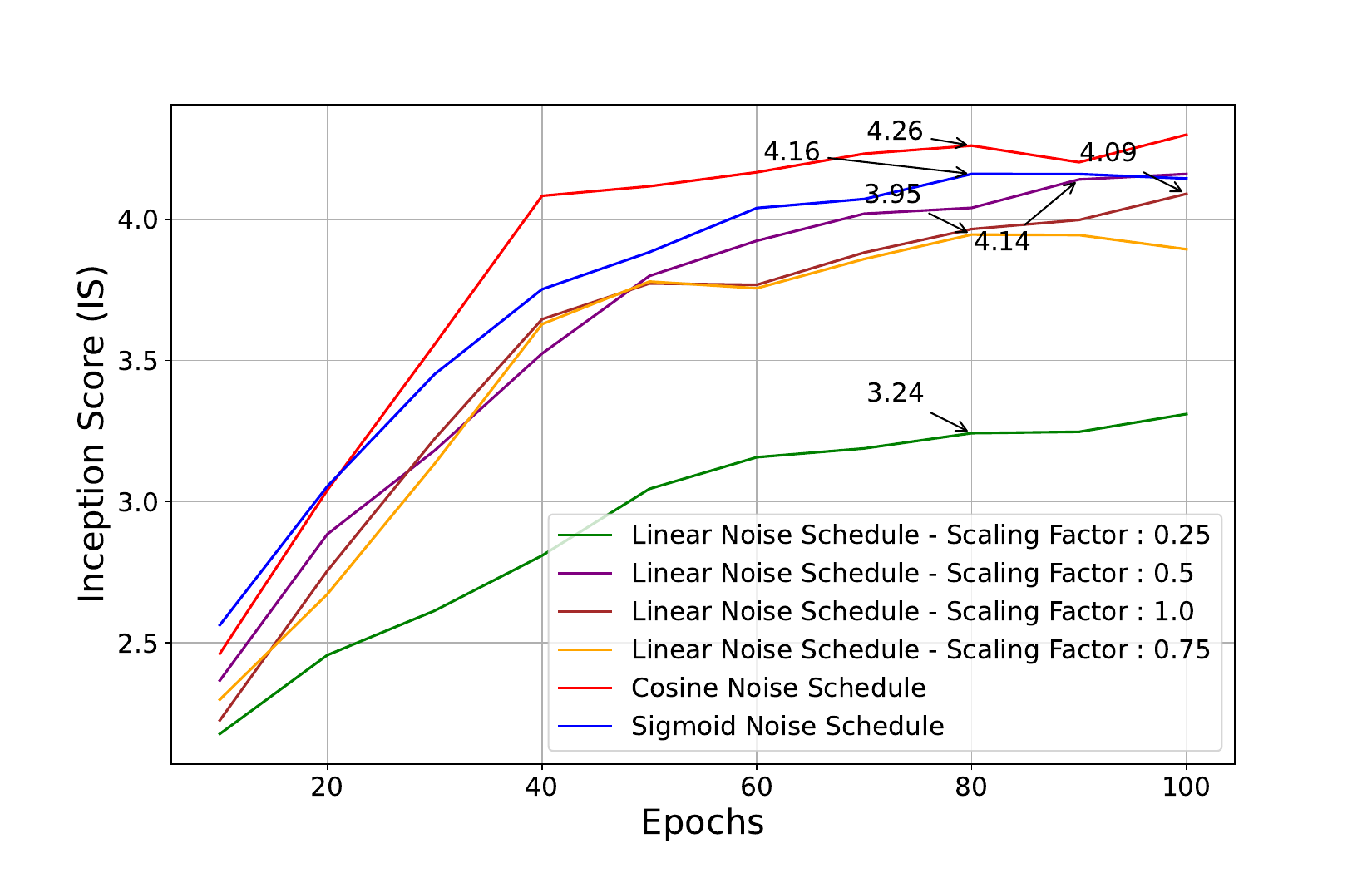}
        \subcaption{}
        \label{fig:Misguided_2}
    \end{subfigure}
    \end{center}
    \end{adjustwidth}
    \caption{IS as a function of epochs of DDPM trained on CIFAR10, using different noise schedules and sampled with the Markovian sampler  under a) Classifier Guidance and b) Misguided Diffusion.}
    \label{fig:markovian_misguided}
\end{figure}

In addition to our findings that show the behavior of the DDIM sampler under Classifier Guidance and Misguided Diffusion in Figure~\ref{fig:IS_Guidance}, the above results with the Markovian sampler, give rise to the question of how the samplers behave when the classifier gradients are incorporated into the sampling process regardless of the state of the classifier (pretrained or untrained). The above experimental findings show that incorporating the classifier gradient into the sampling process has little affect on the sampling process. Classifier Guidance is beneficial for generating class-homogeneous images, but the above results show that it does not universally enhance image quality. Instead, the inherent capabilities of the DMs, powered by their architectural design and diffusion mechanics, are the primary drivers of their generative success. 

\subsection{Sampling with ODE Numerical Solvers}

\vspace{-24pt}

\begin{figure}[H]
    \begin{adjustwidth}{-2cm}{-2cm}
    \begin{center}
    \begin{subfigure}{0.5\textwidth}
        \centering
        \includegraphics[width=\textwidth]{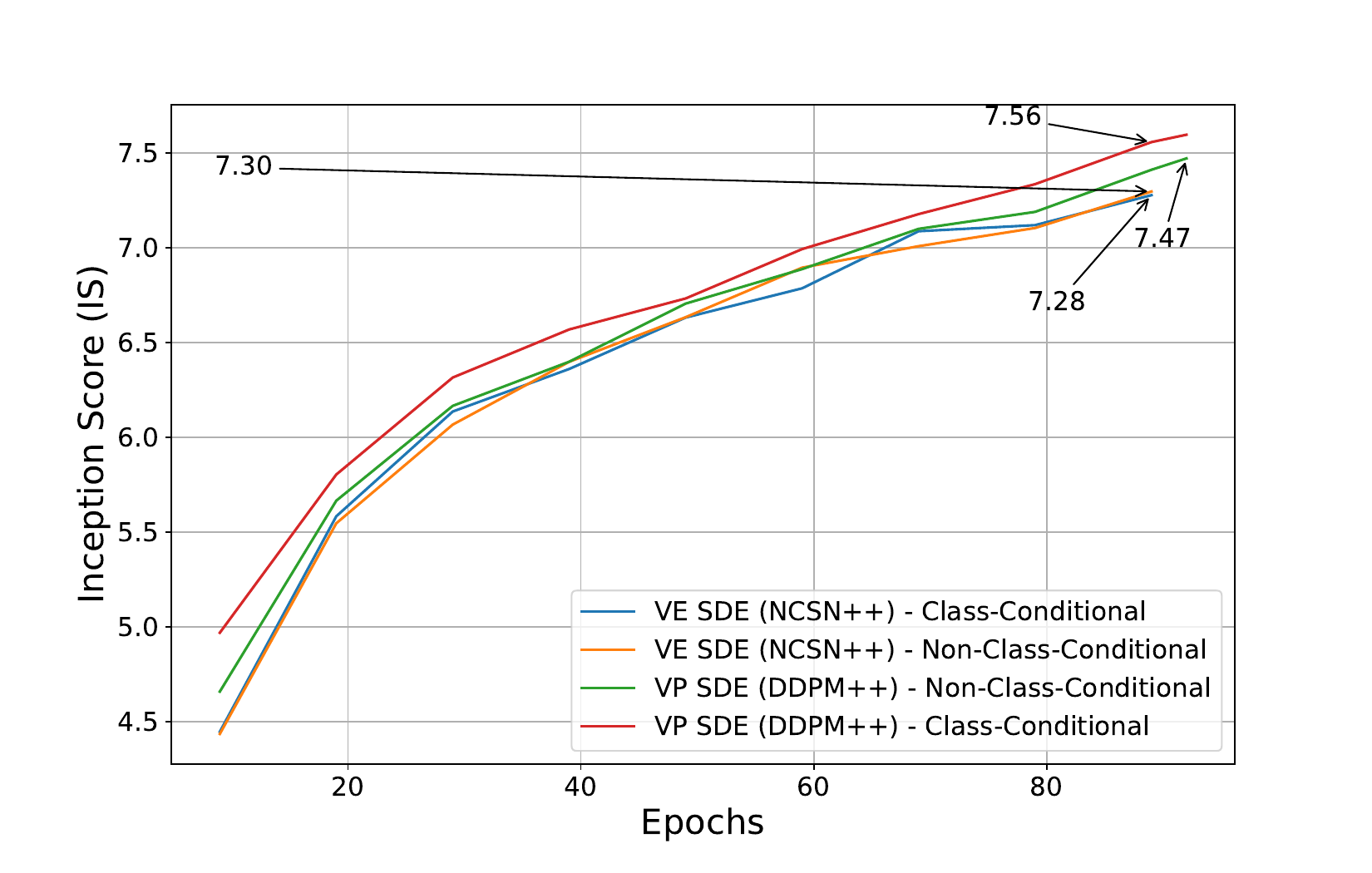}
        \subcaption{}
    \end{subfigure}
    \begin{subfigure}{0.5\textwidth}
        \includegraphics[width=\textwidth]{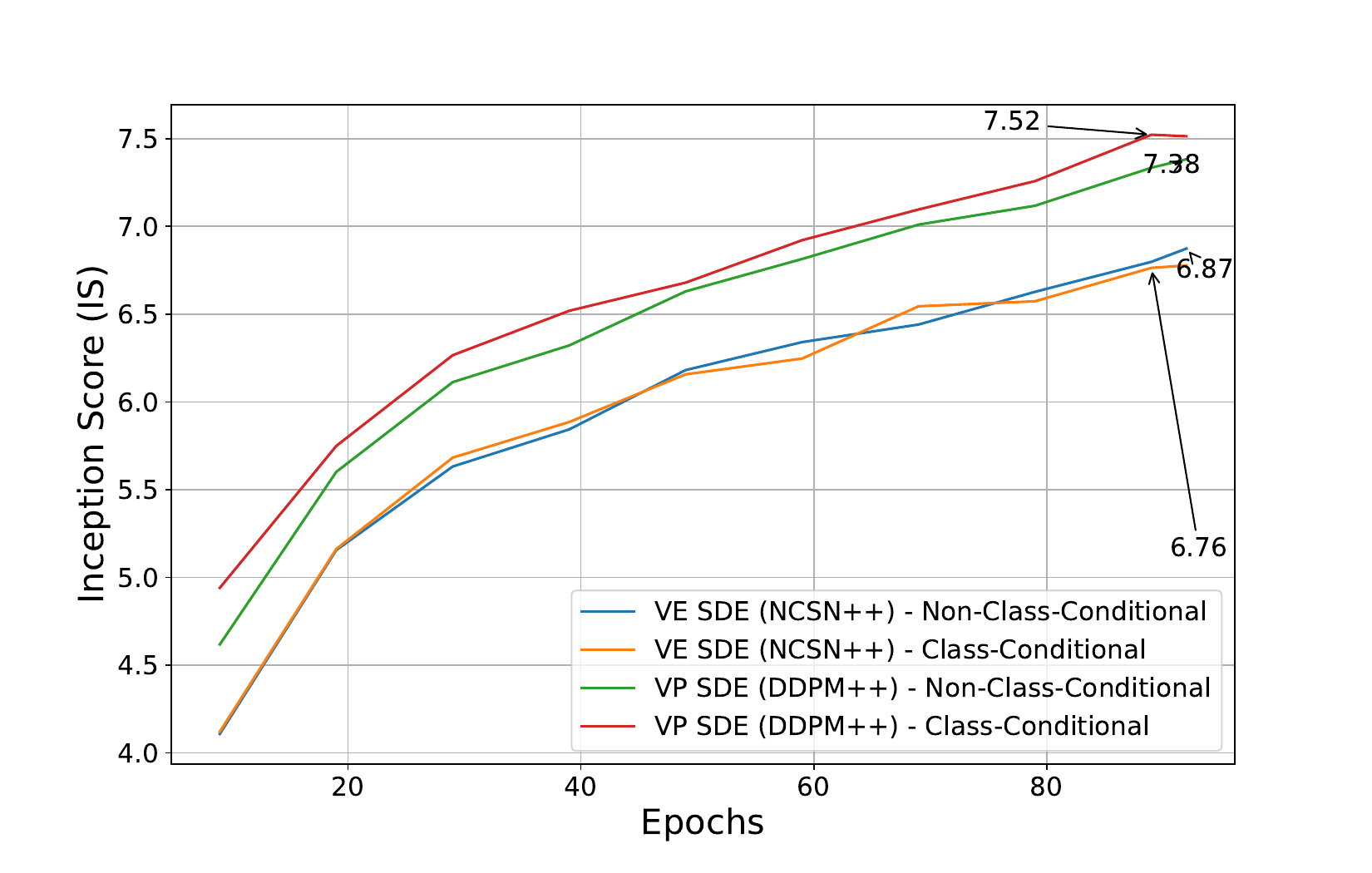}
        \subcaption{}
    \end{subfigure}
    \end{center}
    \end{adjustwidth}
    \caption{ IS as a function of epochs of VE and VP SDE DMs trained on CIFAR10 with and without class conditioning, and sampled using a) the original implementation of Euler's ODE solver and b) the Karras et al. reimplemented Euler's ODE solver (\citeyear{karras2022elucidating}).}
    \label{fig:Eulers_method}
\end{figure}

\begin{figure}[H]
    \begin{adjustwidth}{-2cm}{-2cm}
    \begin{center}
    \begin{subfigure}{0.5\textwidth}
        \centering
        \includegraphics[width=\textwidth]{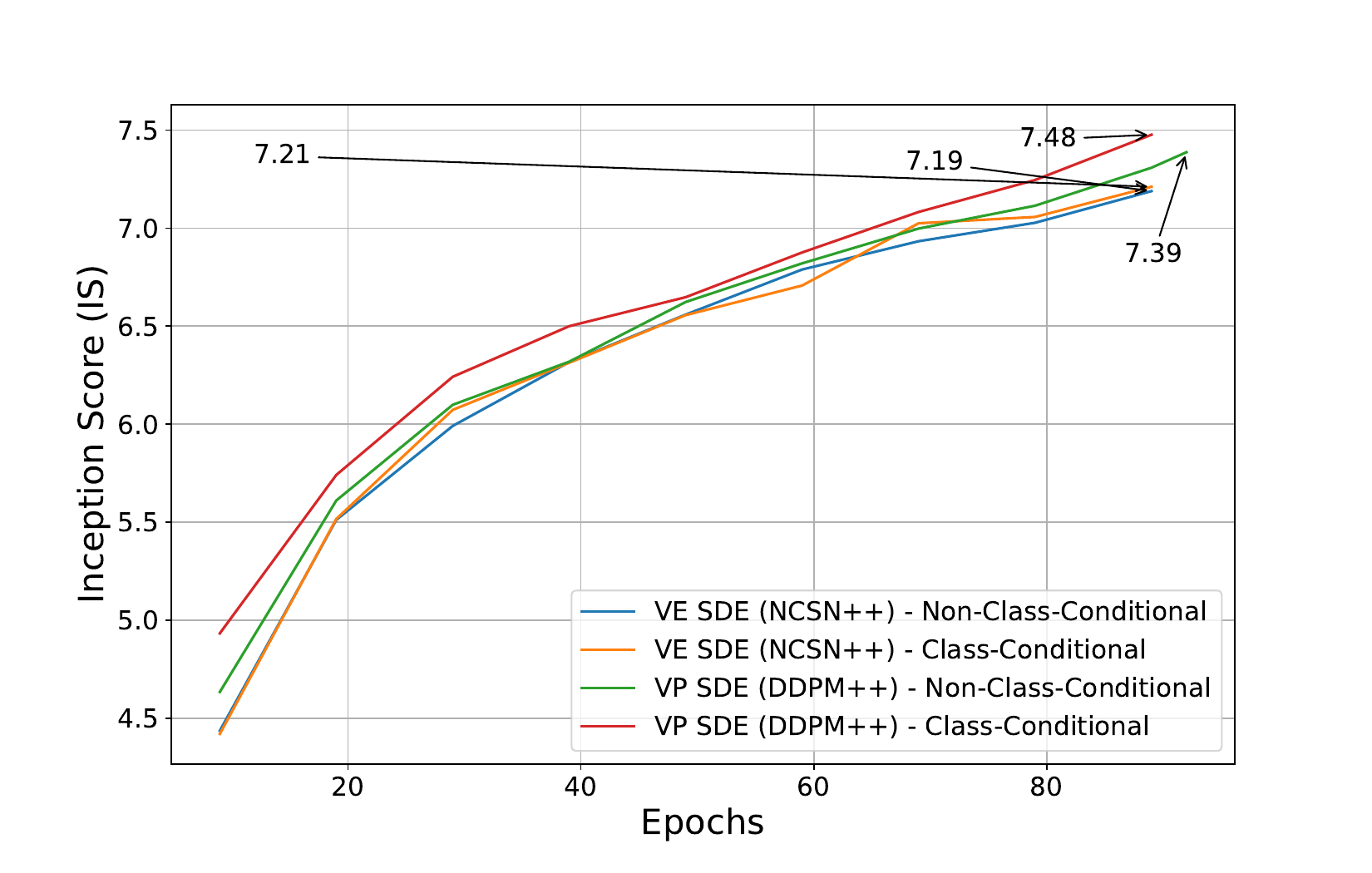}
    \end{subfigure}
    \begin{subfigure}{0.5\textwidth}
        \includegraphics[width=\textwidth]{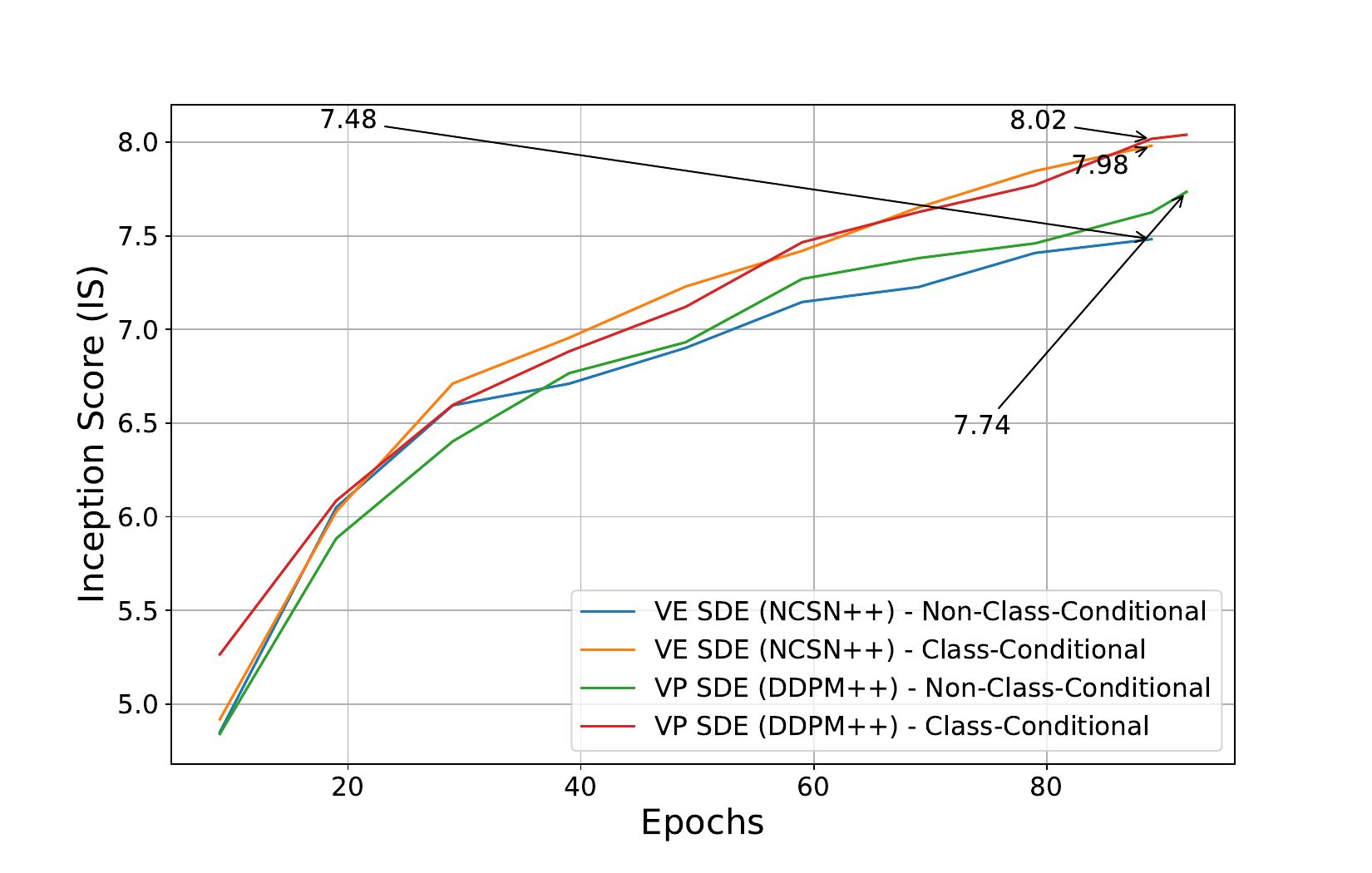}
    \end{subfigure}
    \end{center}
    \end{adjustwidth}
    \caption{  IS as a function of epochs of VE and VP SDE DMs trained on CIFAR10 with and without class conditioning, and sampled using a) the Heuns 2nd Order ODE solver and b) the stochastic sampler.}
    \label{fig:Huens_Stochastic}
\end{figure}

Our experiments aimed to compare the performance of various ODE solvers and the impact of class conditioning on the Inception Score (IS) during the sampling process. As depicted above, we observed that the choice of ODE solver—whether the original implementation of Euler's method, the reimplementation by Karras et al. (\citeyear{karras2022elucidating}), Huen's sampler or the stochastic sampler had a minimal effect on the IS across epochs. This finding suggests that the underlying mechanism of the ODE solver does not significantly influence the sampling performance within the tested domain. Furthermore, the inclusion of class conditioning appeared to have a negligible impact on the IS. This outcome held true across both the VE and VP SDE DM variants, indicating that diffusion guidance does not substantially enhance the IS in our CIFAR10 trained models. A consistent trend across our experiments was the superior performance of the VP SDE DMs over their VE counterparts. Regardless of the presence of class conditioning, VP SDE DMs consistently achieved higher IS scores. This performance discrepancy underscores the effectiveness of DDPM-based diffusion over NCSN-based diffusion. Training the models closer to convergence should reveal even more details about the behavior of each sampler and each trained DM models.

\begin{figure}[H]
    \begin{center}
    \centering
    \begin{subfigure}{0.7\textwidth}
        \includegraphics[width=\textwidth]{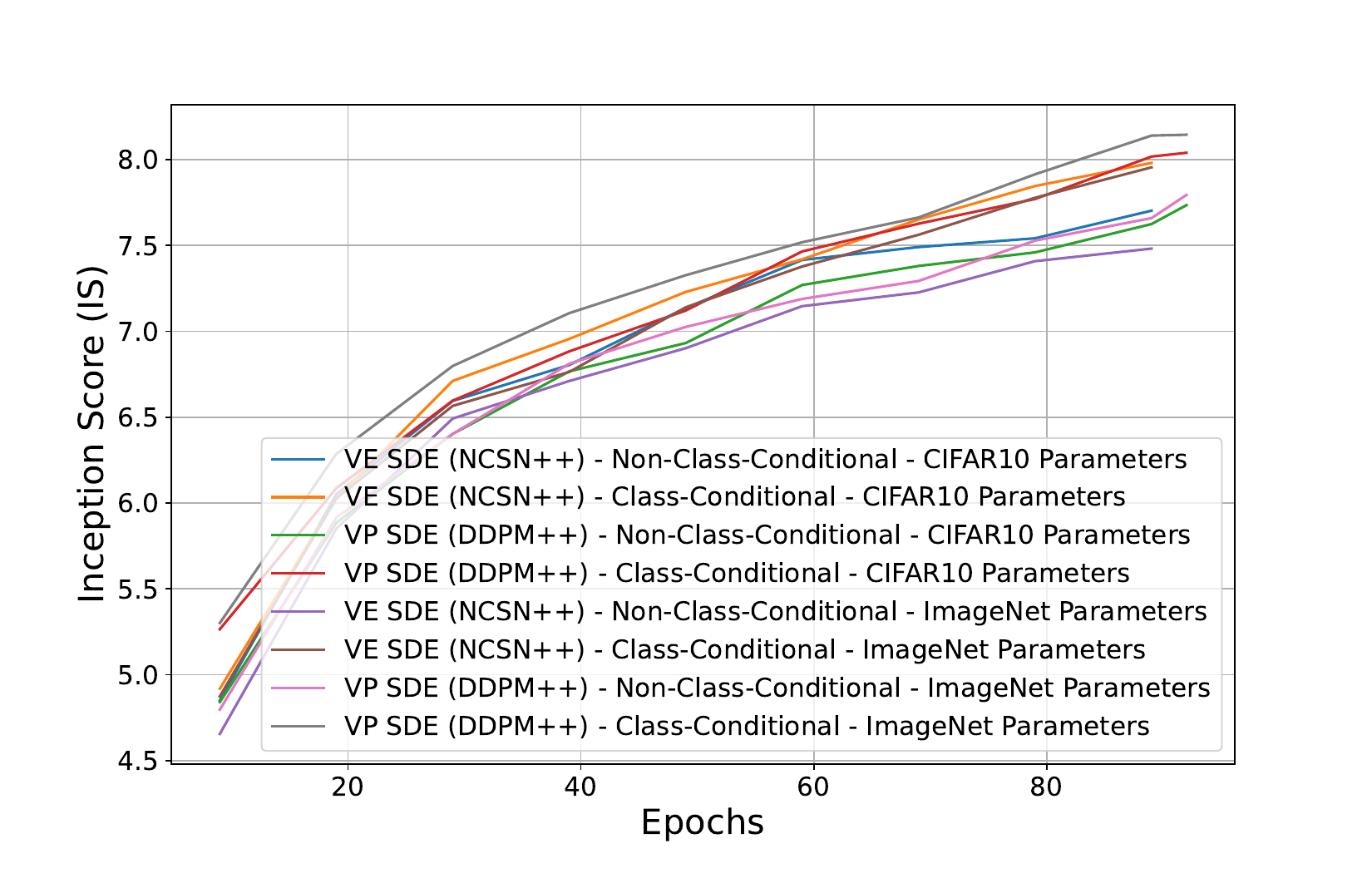 }
    \end{subfigure}

    \caption{IS as a function of epochs of of VE and VP SDE DMs trained on CIFAR10 with and without class conditioning, and sampled using the stochastic sampler using "ImageNet parameters" and "CIFAR10 parameters".}
    \label{fig:Stochastic_parameters}
    \end{center}
\end{figure}

Figure~\ref{fig:Stochastic_parameters} shows the stochastic sampler being used with ImageNet parameters and CIFAR10 parameters being used on models trained on CIFAR10. The ImageNet parameters take the least computational time, as shown in Tables~\ref{table:experiment_times} and \ref{table:experiment_times_2}, and yield results that are within the same IS range as the CIFAR10 parameters.

\subsection{Instability of Image Generation with NCSN}

\begin{figure}[H]
    \centering
    \begin{subfigure}{0.7\textwidth}
        \includegraphics[width=\textwidth]{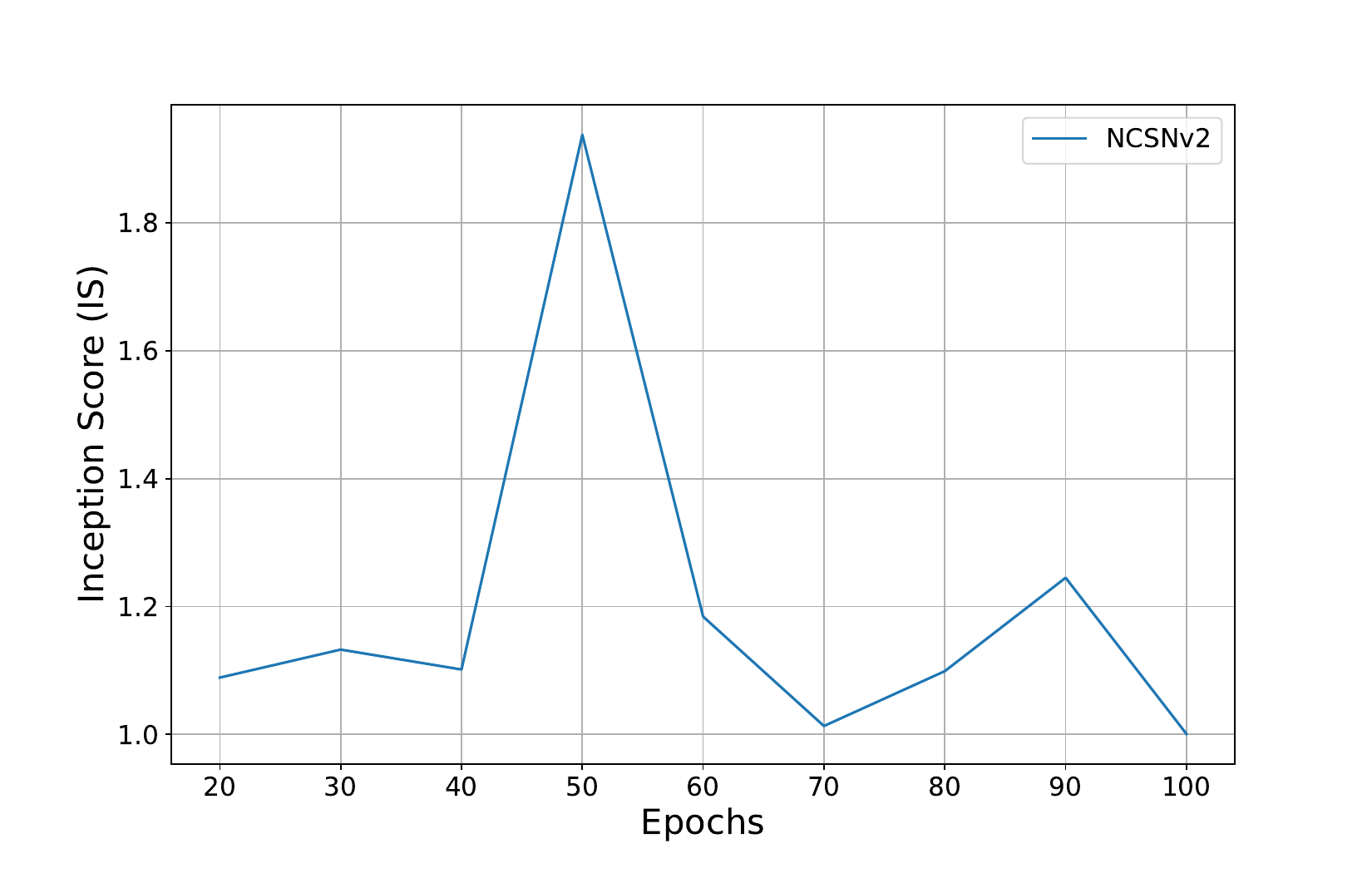}
    \end{subfigure}

    \caption{\label{fig:NCSNv2}  IS as a function of epochs of NCSNv2 sampled with Langevin dynamics}
    
\end{figure}

The traditional discretized NCSN model is too unstable to achieve convergence, as shown by our results in Figure~\ref{fig:NCSNv2} when sampling from a trained NCSNv2. The images stayed noisy without taking any shape or form at any training epoch. The increase in IS at 50 epochs is due to the images taking on a less noisy form. The images then reverted to becoming noisy again after 50 epochs.

\subsection{CIFAR10 Samples}

\begin{figure}[H]
    \begin{center}
    \begin{subfigure}{0.7\textwidth}
        \centering
        \includegraphics[width=\textwidth]{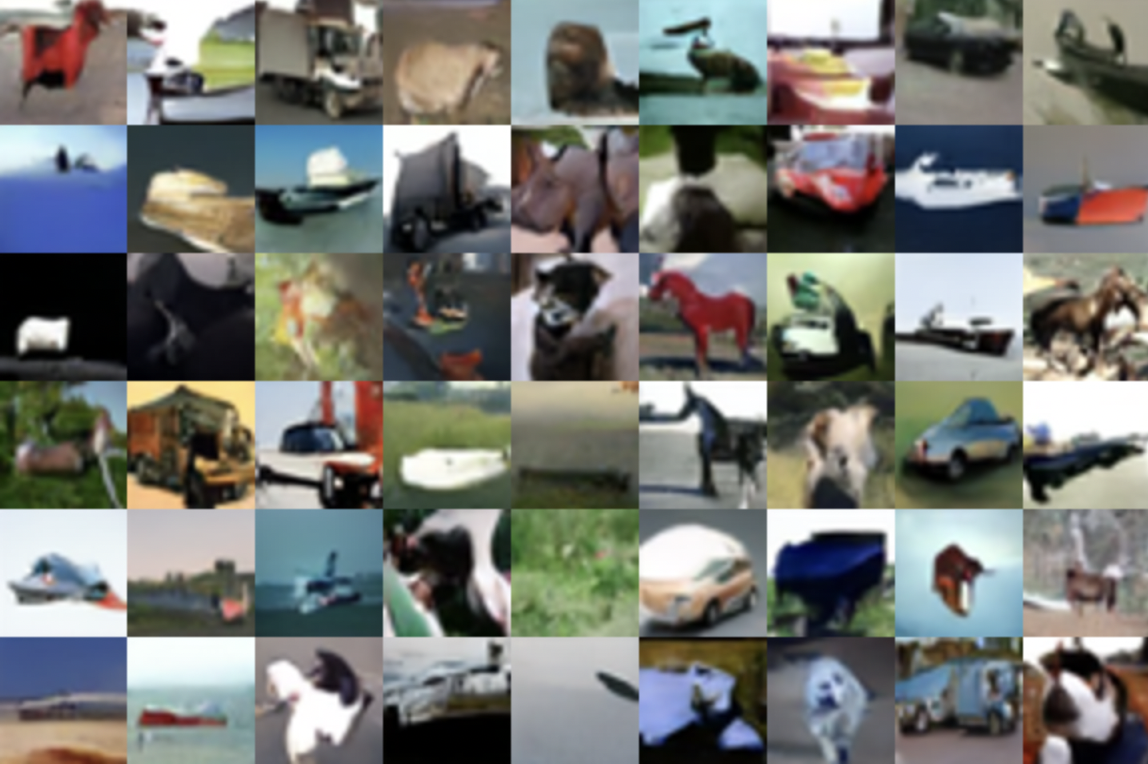}
    \end{subfigure}
    \hspace{0.001\textwidth}
    \caption{ DDPM with cosine noise schedule and Markovian sampling trained for 80 epochs on CIFAR10.}
    \label{fig:DDPM_CIFAR10_80}
    \end{center}
\end{figure}

\begin{figure}[H]
    \begin{center}
    \begin{subfigure}{0.7\textwidth}
        \centering
        \includegraphics[width=\textwidth]{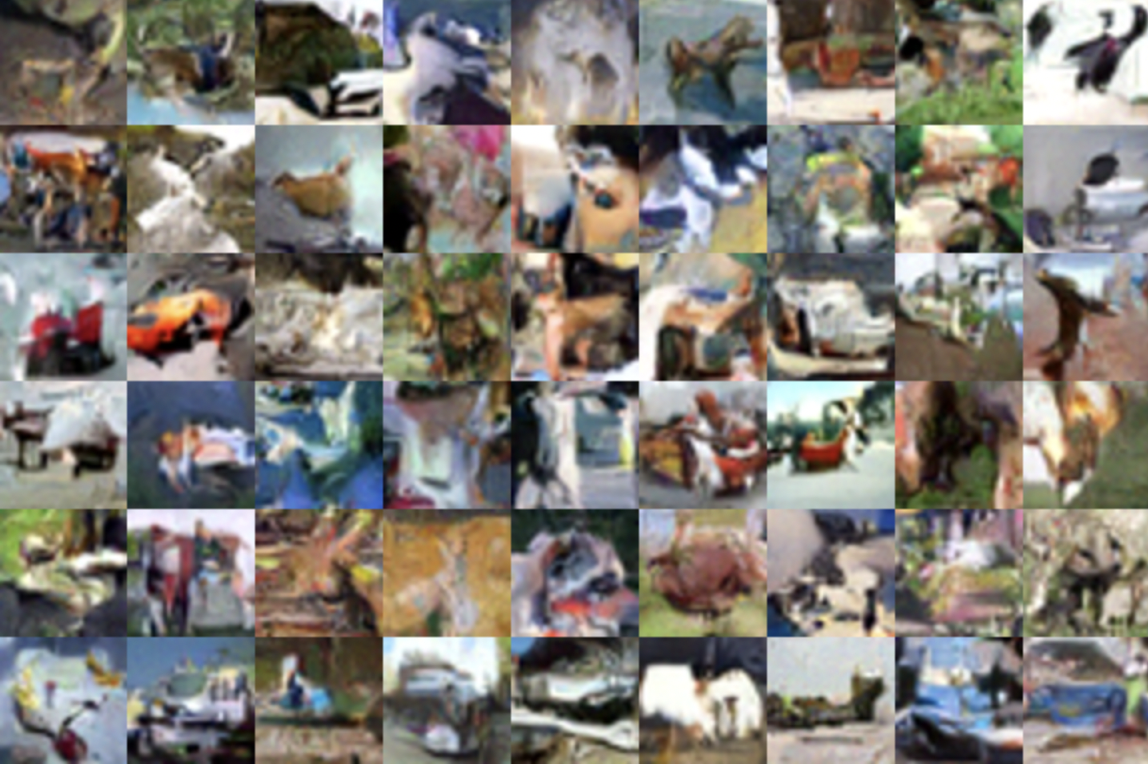}
    \end{subfigure}
    \hspace{0.001\textwidth}
    \caption{ Non-Class Conditional NCSN++ SDE trained for 10 epochs on CIFAR10 and sampled with the Karras et al. reimplemented Euler’s ODE solver (\citeyear{karras2022elucidating}). }
    \label{fig:SDE_NCSN_10_epoch_CIFAR10}
    \end{center}
\end{figure}

\begin{figure}[H]
    \begin{center}
    \begin{subfigure}{0.7\textwidth}
        \centering
        \includegraphics[width=\textwidth]{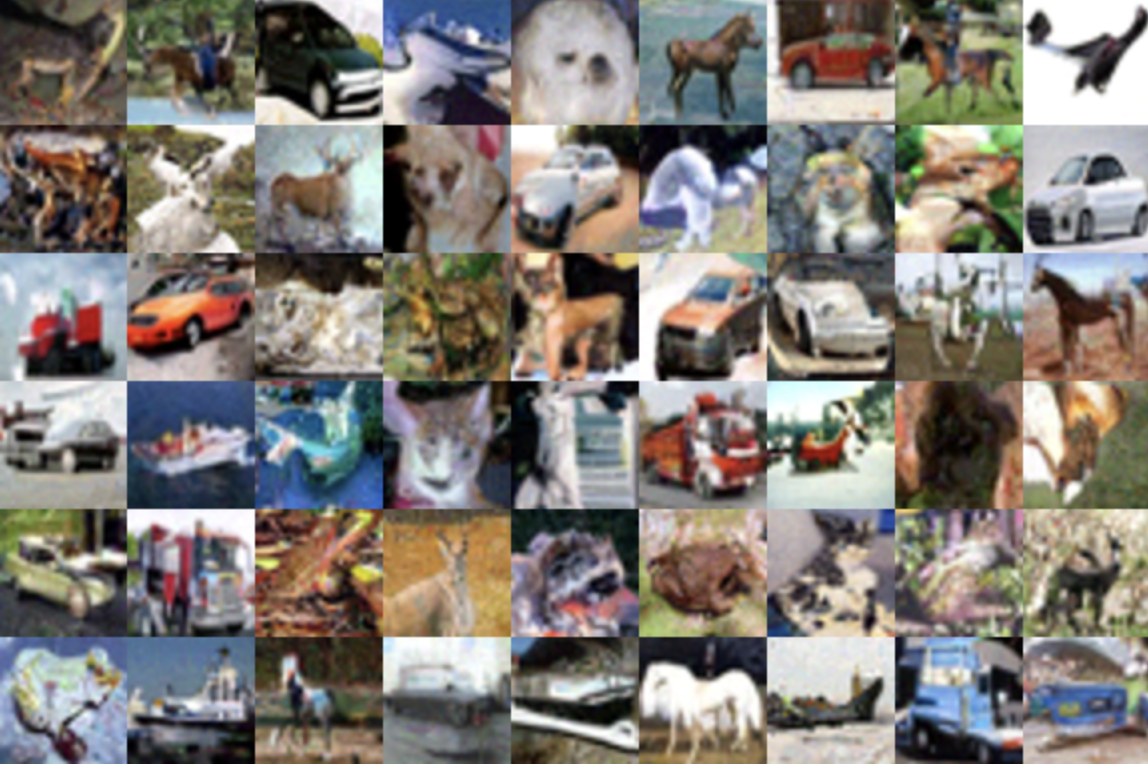}
    \end{subfigure}
    \hspace{0.001\textwidth}
    \caption{ Non-Class Conditional NCSN++ SDE trained for 100 epochs on CIFAR10 and sampled with the Karras et al. reimplemented Euler’s ODE solver (\citeyear{karras2022elucidating}).}
    \label{fig:SDE_NCSN_100_epoch_CIFAR10}
    \end{center}
\end{figure}

\subsection{FFHQ/ImageNet Samples}

\begin{figure}[H]
    \begin{center}
    \begin{subfigure}{0.7\textwidth}
        \centering
        \includegraphics[width=\textwidth]{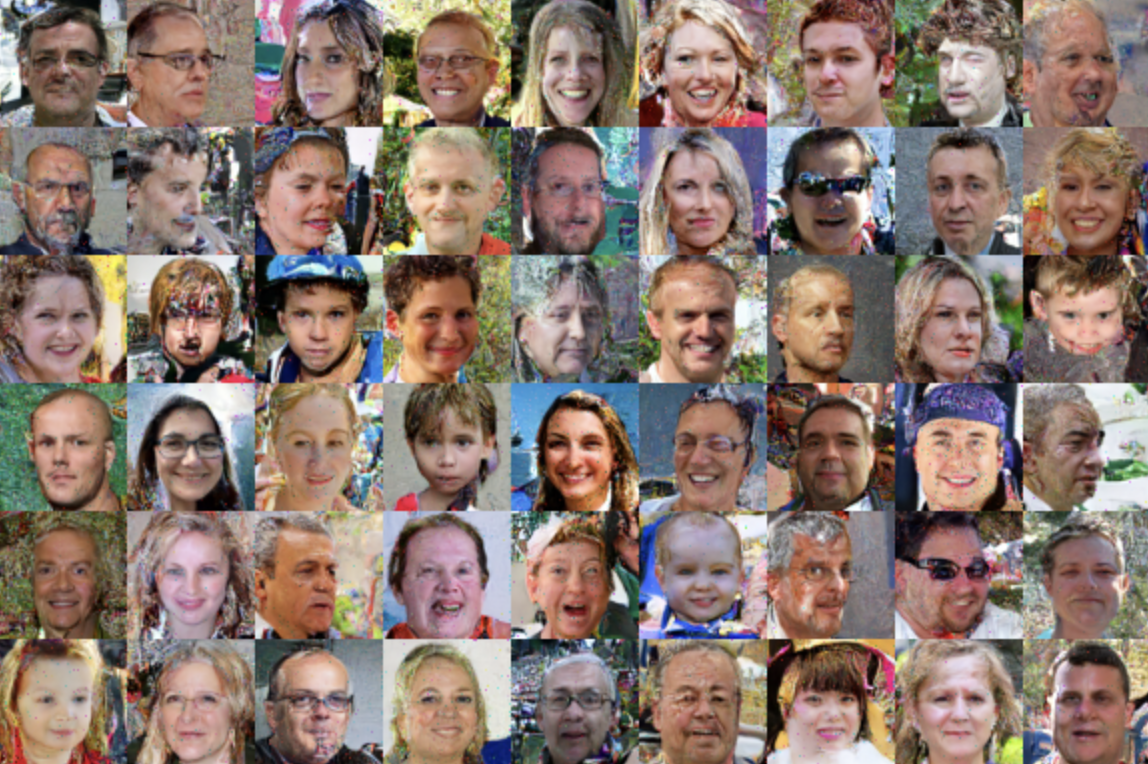}
    \end{subfigure}
    \hspace{0.001\textwidth}
    \caption{DDPM with cosine noise schedule and Markovian sampling trained for 80 epochs on FFHQ.}
    \label{fig:DDPM_FFHQ_80}
    \end{center}
\end{figure}

\begin{figure}[H]
    \begin{center}
    \begin{subfigure}{0.7\textwidth}
        \centering
        \includegraphics[width=\textwidth]{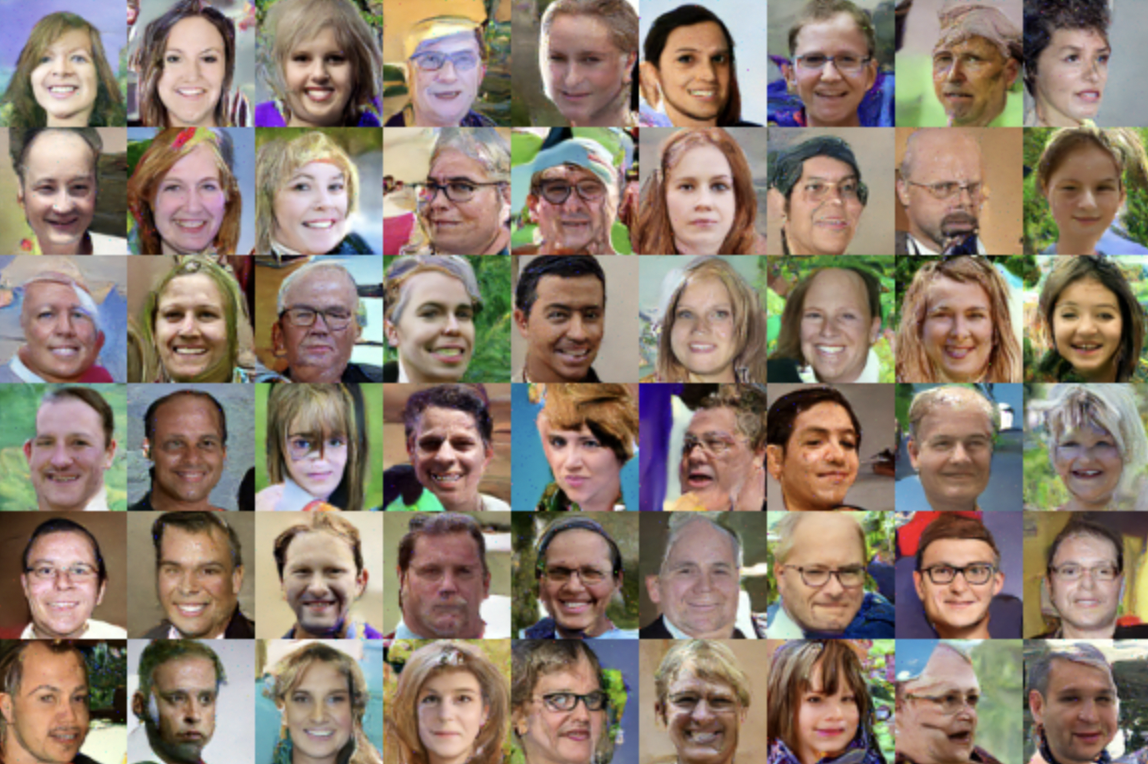}
    \end{subfigure}
    \hspace{0.001\textwidth}
    \caption{DDPM with cosine noise schedule and Markovian sampling trained for 150 epochs on FFHQ.}
    \label{fig:DDPM_FFHQ_150}
    \end{center}
\end{figure}

\begin{figure}[H]
    \begin{center}
    \begin{subfigure}{0.7\textwidth}
        \centering
        \includegraphics[width=\textwidth]{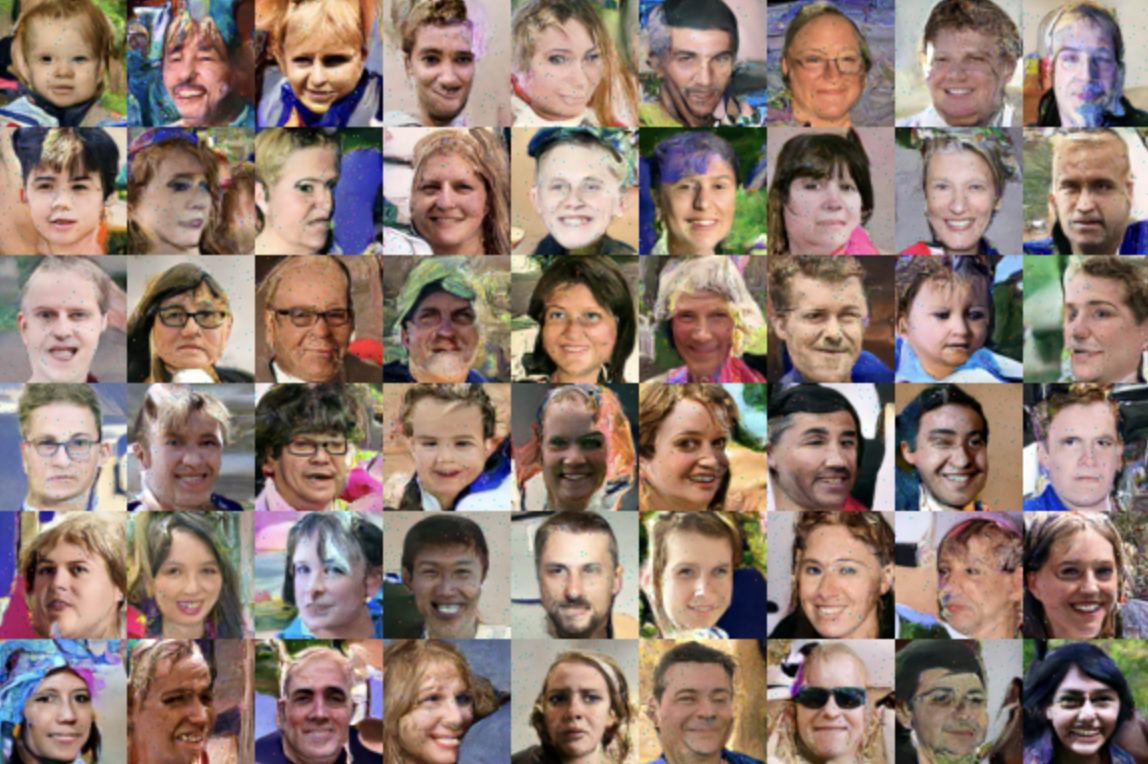}
    \end{subfigure}
    \hspace{0.001\textwidth}
    \caption{DDPM with cosine noise schedule and Markovian sampling trained for 200 epochs on FFHQ.}
    \label{fig:DDPM_FFHQ_200}
    \end{center}
\end{figure}

\begin{figure}[H]
    \begin{center}
    \begin{subfigure}{0.7\textwidth}
        \centering
        \includegraphics[width=\textwidth]{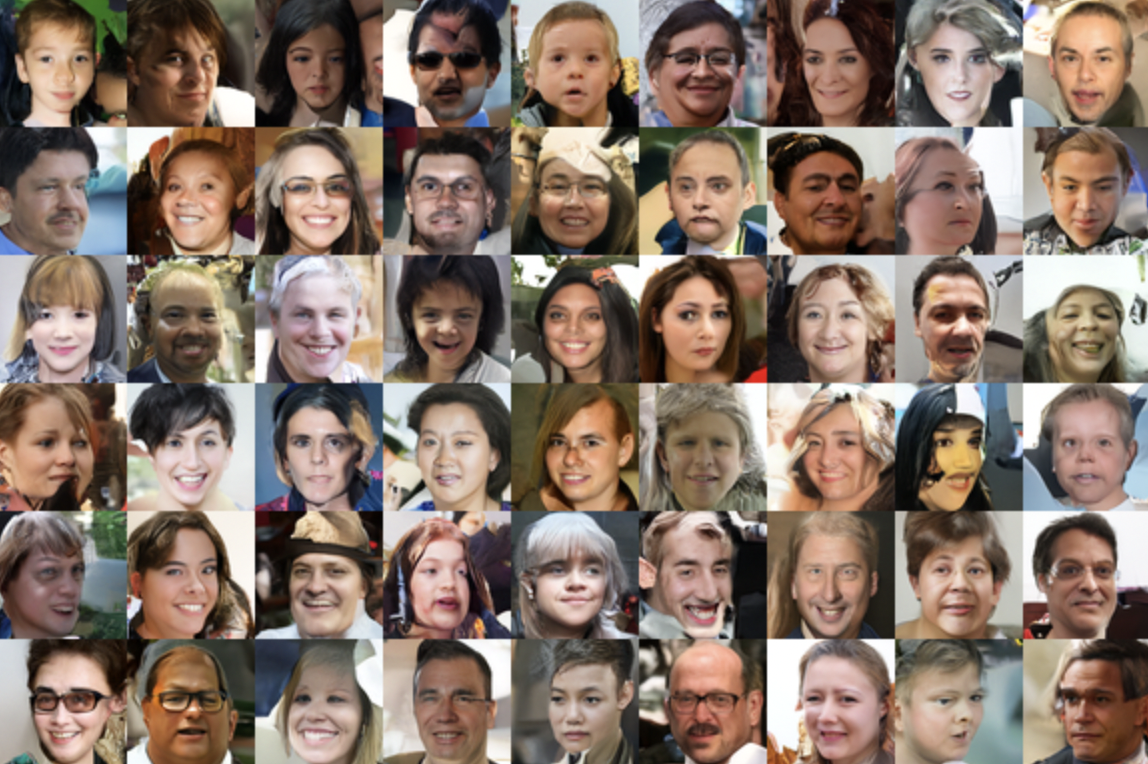}
    \end{subfigure}
    \hspace{0.001\textwidth}
    \caption{Non-Class Conditional DDPM++ SDE trained for 20 epochs on FFHQ and sampled with the Karras et al. reimplemented Euler’s ODE solver (\citeyear{karras2022elucidating}).}
    \label{fig:SDE_DDPM_FFHQ_20}
    \end{center}
\end{figure}

\begin{figure}[H]
    \begin{center}
    \begin{subfigure}{0.7\textwidth}
        \centering
        \includegraphics[width=\textwidth]{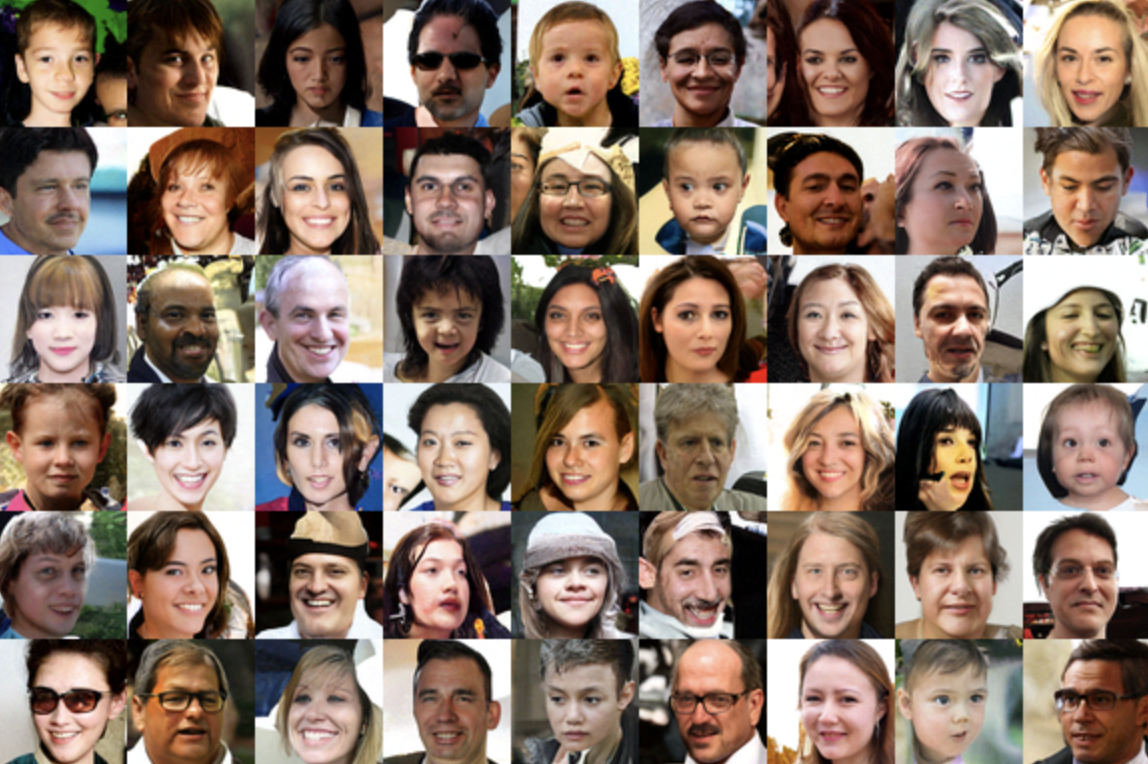}
    \end{subfigure}
    \hspace{0.001\textwidth}
    \caption{Non-Class Conditional DDPM++ SDE trained for 100 epochs on FFHQ and sampled with the Karras et al. reimplemented Euler’s ODE solver (\citeyear{karras2022elucidating}).}
    \label{fig:SDE_DDPM_FFHQ_100}
    \end{center}
\end{figure}

\begin{figure}[H]
    \begin{center}
    \begin{subfigure}{0.7\textwidth}
        \centering
        \includegraphics[width=\textwidth]{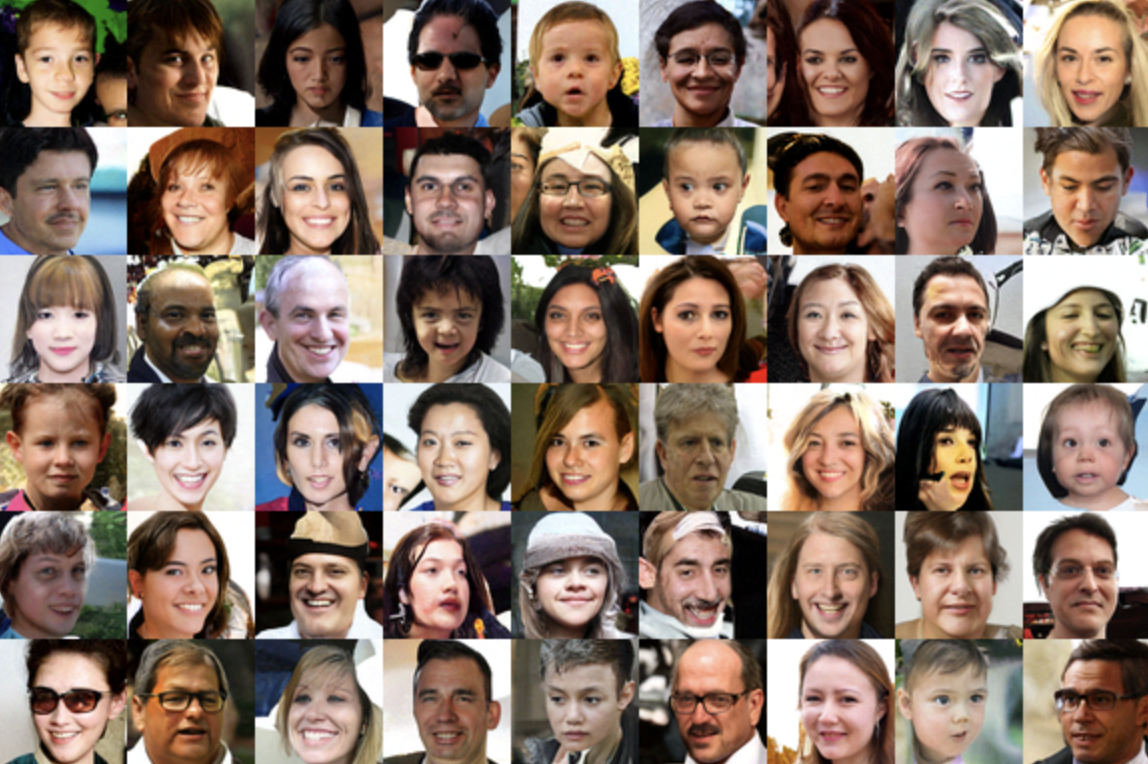}
    \end{subfigure}
    \hspace{0.001\textwidth}
    \caption{Non-Class Conditional NCSN++ SDE trained for 100 epochs on FFHQ and sampled with the Karras et al. reimplemented Euler’s ODE solver (\citeyear{karras2022elucidating}).}
    \label{fig:SDE_NCSN_FFHQ_100}
    \end{center}
\end{figure}

\begin{figure}[H]
    \begin{center}
    \begin{subfigure}{0.7\textwidth}
        \centering
        \includegraphics[width=\textwidth]{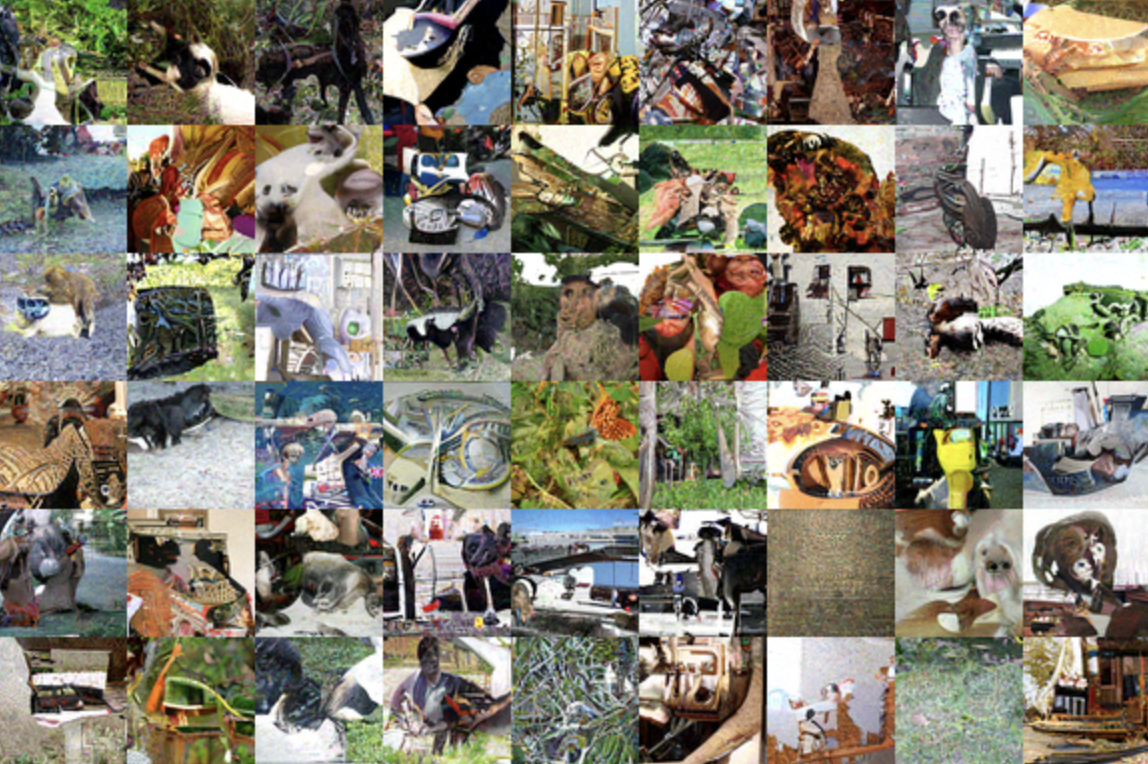}
    \end{subfigure}
    \hspace{0.001\textwidth}
    \caption{Non-Class Conditional NCSN++ SDE trained for 5 epochs on ImageNet and sampled with the Karras et al. reimplemented Euler’s ODE solver (\citeyear{karras2022elucidating}).}
    \label{fig:SDE_NCSN_IMAGENET_100}
    \end{center}
\end{figure}

\end{document}